\documentclass[10pt,twocolumn,letterpaper]{article}

\usepackage{cvpr}
\usepackage{times}
\usepackage{epsfig}
\usepackage{graphicx}
\usepackage{amsmath}
\usepackage{amssymb}

\usepackage{booktabs}
\usepackage{xspace}
\usepackage{multirow}
\usepackage[dvipsnames]{xcolor}
\usepackage{footmisc}
\usepackage{soul} \usepackage{chngcntr}

\newcommand{\methodfullname}{Multimodal Multi-Copy Mesh\xspace}
\newcommand{\methodname}{M4C\xspace}
\newcommand{\myparagraph}[1]{\vspace{-1.4em}\paragraph{#1.}}
\newcommand{\myparagraphfirst}[1]{\paragraph{#1.}}

\newcommand{\vocab}[1]{\textcolor{Cyan}{\textbf{#1}}}
\newcommand{\ocr}[1]{\textcolor{Dandelion}{\textbf{#1}}}
\newcommand{\figmargin}{-1.25em}
\newcommand{\figcapmargin}{0.05em}
\newcommand{\tabmargin}{-1.25em}
\newcommand{\tabcapmargin}{-0.7em}
\newcommand{\doublecoltopmargin}{-1.5em}

\usepackage[pagebackref=true,breaklinks=true,colorlinks,bookmarks=false]{hyperref}

\cvprfinalcopy 
\def\cvprPaperID{****} 

\begin{document}

\title{Iterative Answer Prediction with Pointer-Augmented\\Multimodal Transformers for TextVQA}

\author{Ronghang Hu$^{1,2}$ $\qquad$ Amanpreet Singh$^{1}$ $\qquad$ Trevor Darrell$^{2}$ $\qquad$ Marcus Rohrbach$^{1}$ \\
$^1$Facebook AI Research (FAIR) $\qquad$ $^2$University of California, Berkeley \\
{\tt\small \{ronghang,trevor\}@eecs.berkeley.edu, \{asg,mrf\}@fb.com}}
\maketitle

\begin{abstract}
\vspace{-0.5em}
Many visual scenes contain text that carries crucial information, and it is thus essential to understand text in images for downstream reasoning tasks. For example, a \emph{deep water} label on a warning sign warns people about the danger in the scene. Recent work has explored the TextVQA task that requires reading and understanding text in images to answer a question. However, existing approaches for TextVQA are mostly based on custom pairwise fusion mechanisms 
between a pair of two modalities and are restricted to a single prediction step by casting TextVQA as a classification task. In this work, we propose a novel model for the TextVQA task based on a multimodal transformer architecture accompanied by a rich representation for text in images. Our model naturally fuses different modalities homogeneously by embedding them into a common semantic space where self-attention is applied to model inter- and intra- modality context. Furthermore, it enables iterative answer decoding with a dynamic pointer network, allowing the model to form an answer through multi-step prediction instead of one-step classification. Our model outperforms existing approaches on three benchmark datasets for the TextVQA task by a large margin.
\end{abstract}
\vspace{-1.2em}

\section{Introduction}
\label{sec:introduction}

\begin{figure}[t]
\centering
\includegraphics[width=0.95\linewidth,trim=0 227 0 0,clip=true]{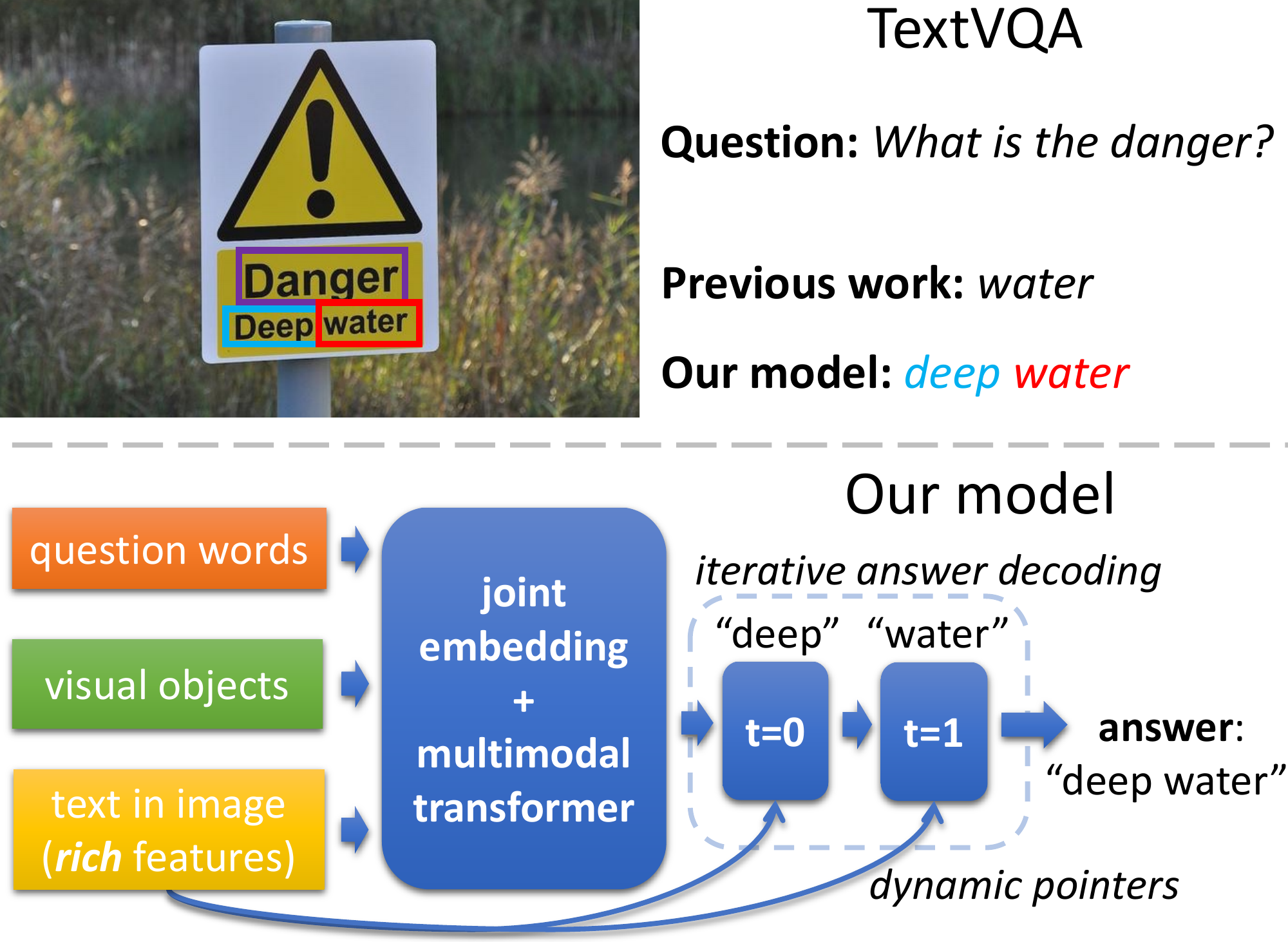}
\caption{Compared to previous work (\eg \cite{singh2019towards}) on the TextVQA task, our model, accompanied by rich features for image text, handles all modalities with a multimodal transformer over a joint embedding space instead of pairwise fusion mechanisms between modalities. Furthermore, answers are predicted through iterative decoding with pointers instead of one-step classification over a fixed vocabulary or copying single text token from the image.}
\label{fig:teaser}
\vspace{\figmargin}
\end{figure}

As a prominent task for visual reasoning, the Visual Question Answering (VQA) task \cite{antol2015vqa} has received wide attention in terms of both datasets (\eg \cite{antol2015vqa,goyal2017making,kafle2017analysis,johnson2017clevr,hudson2019gqa}) and methods (\eg \cite{fukui2016multimodal,anderson2018bottom,ben2017mutan,kim2018bilinear,lu2019vilbert}). However, these datasets and methods mostly focus on the visual components in the scene. On the other hand, they tend to ignore a crucial modality -- text in the images -- that carries essential information for scene understanding and reasoning. For example, in Figure~\ref{fig:teaser}, \textit{deep water} on the sign warns people about the danger in the scene. 
To address this drawback, new VQA datasets \cite{singh2019towards,biten2019scene,mishraocr19} have been recently proposed with questions that explicitly require understanding and reasoning about text in the image, which is referred to as the TextVQA task. 

The TextVQA task distinctively requires models to see, read and reason over three modalities: the input question, the visual contents in the image such as visual objects, and the text in the image. Several approaches \cite{singh2019towards,biten2019scene,mishraocr19,biten2019icdar} have been proposed for the TextVQA task, based on OCR results of the image. In particular, LoRRA \cite{singh2019towards} extends previous VQA models \cite{singh2018pythia} with an OCR attention branch and adds OCR tokens as a dynamic vocabulary to the answer classifier, allowing copying a single OCR token from the image as the answer. Similarly in \cite{mishraocr19}, OCR tokens are grouped into blocks and added to the output space of a VQA model. 

While these approaches enable reading text in images to some extent, they typically rely on custom pairwise multimodal fusion mechanisms between two modalities (such as single-hop attention over image regions and text tokens, conditioned on the input question), which limit the types of possible interactions between modalities. Furthermore, they treat answer prediction as a single-step classification problem -- either selecting an answer from the training set answers or copying a text token from the image -- making it difficult to generate complex answers such as book titles or signboard names with multiple words, or answers with both common words and specific image text tokens, such as \textit{McDonald's burger} where \textit{McDonald's} is from text in the image and \textit{burger} is from the model's own vocabulary. In addition, the word embedding based image text features in previous work have limited representation power and miss important cues such as the appearance (\eg font and color) and the location of text tokens in images. For example, tokens that have different fonts and are spatially apart from each other usually do not belong to the same street sign.

In this paper, we address the above limitations with our novel \methodfullname (\methodname) model for the TextVQA task, based on the transformer \cite{vaswani2017attention} architecture accompanied by iterative answer decoding through dynamic pointers, as shown in Figure~\ref{fig:teaser}. Our model naturally fuses the three input modalities and captures intra- and inter- modality interactions homogeneously within a multimodal transformer, which projects all entities from each modality into a common semantic embedding space, and applies the self-attention mechanism \cite{parikh2016decomposable,vaswani2017attention} to collect relational representations for each entity. Instead of casting answer prediction as a classification task, we perform iterative answer decoding in multiple steps and augment our answer decoder with a dynamic pointer network that allows selecting text in the image in a permutation-invariant way without relying on any ad-hoc position indices in previous work such as LoRRA \cite{singh2019towards}. Furthermore, our model is capable of combining its own vocabulary with text in the image in a generated answer, as shown in examples in Figure~\ref{fig:vis_textvqa} and \ref{fig:vis_stvqa}. Finally, we introduce a rich representation for text tokens in the images based on multiple cues, including its word embedding, appearance, location, and character-level information.

Our contributions in this paper are as follows: 1) We show that multiple (more than two) input modalities can be naturally fused and jointly modeled through our multimodal transformer architecture. 2) Unlike previous work on TextVQA, our model reasons about the answer beyond a single classification step and predicts it through our pointer-augmented multi-step decoder. 3) We adopt a rich feature representation for text tokens in images and show that it is better than features based only on word embedding in previous work. 4) Our model significantly outperforms previous work on three challenging datasets for the TextVQA task: TextVQA \cite{singh2019towards} (+25\% relative), ST-VQA \cite{biten2019scene} (+65\% relative), and OCR-VQA \cite{mishraocr19} (+32\% relative).

\section{Related work}
\label{sec:related_work}

\myparagraphfirst{VQA based on reading and understanding image text} 
Recently, a few datasets and methods \cite{singh2019towards,biten2019scene,mishraocr19,biten2019icdar} have been proposed for visual question answering based on text in images (referred to as the TextVQA task). LoRRA \cite{singh2019towards}, a prominent prior work on this task, extends the Pythia \cite{singh2018pythia} framework for VQA and allows it to copy a single OCR token from the image as the answer, by applying a single attention hop (conditioned on the question) over the OCR tokens and including the OCR token indices in the answer classifier's output space. A conceptually similar model is proposed in \cite{mishraocr19}, where OCR tokens are grouped into blocks and added to both the input features and the output answer space of a VQA model. In addition, a few other approaches \cite{biten2019scene,biten2019icdar} enable text reading by augmenting existing VQA models with OCR inputs. However, these existing methods are limited by their simple feature representation of image text, multimodal learning approaches, and one-step classification for answer outputs. In this work, we address these limitations with our \methodname model.

\myparagraph{Multimodal learning in vision-and-language tasks} Early approaches on vision-and-language tasks often combined the image and text through attention over one modality conditioned on the other modality, such as image attention based on text (\eg \cite{yang2016stacked,lu2016hierarchical}). Some approaches have explored multimodal fusion mechanisms such as bilinear models (\eg \cite{fukui2016multimodal,kim2018bilinear}), self-attention (\eg \cite{gao2019dynamic}), and graph networks (\eg \cite{li2019relation}). Inspired by the success of Transformer \cite{vaswani2017attention} and BERT \cite{devlin2019bert} architectures in natural language tasks, several recent works \cite{lu2019vilbert,alberti2019fusion,tan2019lxmert,li2019visualbert,li2019unicoder,su2019vl,zhou2019unified,chen2019uniter} have also applied transformer-based fusion between image and text with self-supervision on large-scale datasets. However, most existing works treat each modality with a specific set of parameters, which makes them hard to scale to more input modalities. On the other hand, in our work we project all entities from each modality into a joint embedding space and treat them homogeneously with a transformer architecture over the list of all things. Our results suggest that joint embedding and self-attention are efficient when modeling multiple (more than two) input modalities.

\myparagraph{Dynamic copying with pointers}
Many answers in the TextVQA task come from text tokens in the image such as book titles or street signs. As it is intractable to have every possible text token in the answer vocabulary, copying text from the image would often be an easier option for answer prediction. Prior work has explored dynamically copying the inputs in different tasks such as text summarization \cite{see2017get}, knowledge retrieval \cite{yavuz2019deepcopy}, and image captioning \cite{lu2018neural} based on Pointer Networks \cite{vinyals2015pointer} and its variants. For the TextVQA task, recent works \cite{singh2019towards,mishraocr19} have proposed to copy OCR tokens by adding their indices to classifier outputs. However, apart from their limitation of copying only a single token (or block), one drawback of these approaches is that they require a pre-defined number of OCR tokens (since the classifier has a fixed output dimension) and their output is dependent on the ordering of the tokens. In this work, we overcome this drawback using a permutation-invariant pointer network together with our multimodal transformer.

\section{\methodfullname (\methodname)}
\label{sec:method}

\begin{figure*}[t]
\vspace{\doublecoltopmargin}
\centering
\includegraphics[width=0.95\linewidth,trim=0 0 0 0,clip=false]{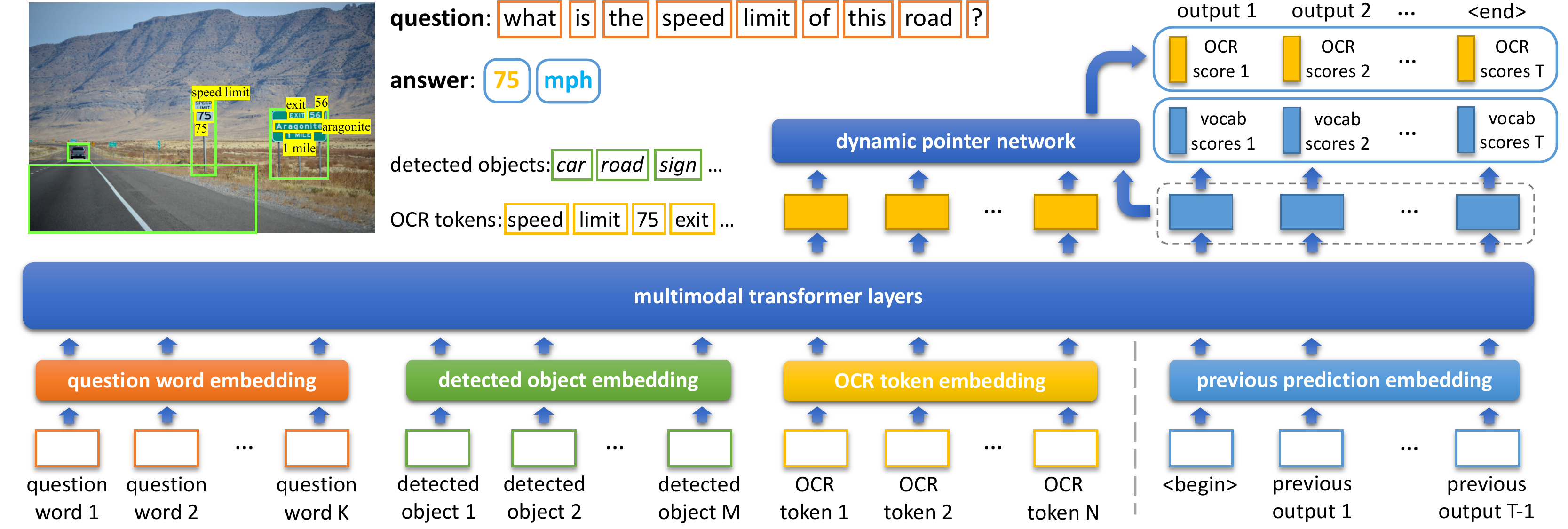}
\caption{An overview of our \methodname model. We project all entities (question words, detected visual objects, and detected OCR tokens) into a common $d$-dimensional semantic space through domain-specific embedding approaches and apply multiple transformer layers over the list of projected things. Based on the transformer outputs, we predict the answer through iterative auto-regressive decoding, where at each step our model either selects an OCR token through our dynamic pointer network, or a word from its fixed answer vocabulary.}
\label{fig:method}
\vspace{\figmargin}
\end{figure*}

In this work, we present \methodfullname (\methodname), a novel approach for the TextVQA task based on a pointer-augmented multimodal transformer architecture with iterative answer prediction. Given a question and an image as inputs, we extract feature representations from three modalities -- the question, the visual objects in the image, and the text present in the image. These three modalities are represented respectively as a list of question words features, a list of visual object features from an off-the-shelf object detector, and a list of OCR token features based on an external OCR system.

Our model projects the feature representations of entities (in our case, question words, detected objects, and detected OCR tokens) from the three modalities as vectors in a learned common embedding space. Then, a multi-layer transformer \cite{vaswani2017attention} is applied on the list of all projected features, enriching their representations with intra- and inter- modality context. Our model learns to predict the answer through iterative decoding accompanied by a dynamic pointer network. During decoding, it feeds in the previous output to predict the next answer component in an auto-regressive manner. At each step, it either copies an OCR token from the image, or selects a word from its fixed answer vocabulary. Figure \ref{fig:method} shows an overview of our model.

\subsection{A common embedding space for all modalities}
\label{sec:method_embedding}

Our model receives inputs from three modalities -- question words, visual objects, and OCR tokens. We extract feature representations for each modality and project them into a common $d$-dimensional semantic space through domain-specific embedding approaches as follows.

\myparagraph{Embedding of question words} Given a question as a sequence of $K$ words, we embed these words into the corresponding sequence of $d$-dimensional feature vectors $\{x^\text{ques}_k\}$ (where $k=1,\cdots,K)$ using a pretrained BERT model \cite{devlin2019bert}.\footnote{In our implementation, we extract question word features from the first 3 layers of BERT-BASE. We find it sufficient to use its first few layers instead of using all its 12 layers, which saves computation.} During training, the BERT parameters are fine-tuned using the question answering loss. 

\myparagraph{Embedding of detected objects} Given an image, we obtain a set of $M$ visual objects through a pretrained detector (Faster R-CNN \cite{ren2015faster} in our case). Following prior work \cite{anderson2018bottom,singh2018pythia,singh2019towards}, we extract appearance feature $x^\text{fr}_m$ using the detector's output from the $m$-th object (where $m=1,\cdots,M)$. To capture its location in the image, we introduce a 4-dimensional location feature $x^\text{b}_m$ from $m$-th object's relative bounding box coordinates $[x_\text{min}/W_\text{im}, y_\text{min}/H_\text{im}, x_\text{max}/W_\text{im}, y_\text{max}/H_\text{im}]$, where $W_\text{im}$ and $H_\text{im}$ are image width and height respectively. Then, the appearance feature and the location feature are projected into the $d$-dimensional space with two learned linear transforms (where $d$ is the same as in the question word embedding above), and are summed up as the final object embedding $\{x^\text{obj}_m\}$ as
\begin{equation}
    x^\text{obj}_m = \text{LN}(W_1 x^\text{fr}_m) + \text{LN}(W_2 x^\text{b}_m)
\end{equation}
where $W_1$ and $W_2$ are learned projection matrices. $\text{LN}(\cdot)$ is layer normalization \cite{ba2016layer}, added on the output of the linear transforms to ensure that the object embedding has the same scale as the question word embedding. We fine-tune the last layer of the Faster R-CNN detector during training.

\myparagraph{Embedding of OCR tokens with rich representations} Intuitively, to represent text in images, one needs to encode not only its characters, but also its appearance (\eg color, font, and background) and spatial location in the image (\eg words appearing on the top of a book cover are more likely to be book titles). We follow this intuition in our model and use a rich OCR representation consisting of four types of features, which is shown in our experiments to be significantly better than word embedding (such as FastText) alone in prior work \cite{singh2019towards}. After obtaining a set of $N$ OCR tokens in an image through external OCR systems, from the $n$-th token (where $n=1,\cdots,N)$ we extract 1) a 300-dimensional FastText \cite{bojanowski2017enriching} vector $x^\text{ft}_n$, which is a word embedding with sub-word information, 2) an appearance feature $x^\text{fr}_n$ from the same Faster R-CNN detector in the object detection above, extracted via RoI-Pooling on the OCR token's bounding box, 3) a 604-dimensional Pyramidal Histogram of Characters (PHOC) \cite{almazan2014word} vector $x^\text{p}_n$, capturing what characters are present in the token -- this is more robust to OCR errors and can be seen as a coarse character model, and 4) a 4-dimensional location feature $x^\text{b}_n$ based on the OCR token's relative bounding box coordinates $[x_\text{min}/W_\text{im}, y_\text{min}/H_\text{im}, x_\text{max}/W_\text{im}, y_\text{max}/H_\text{im}]$. We linearly project each feature into $d$-dimensional space, and sum them up (after layer normalization) as the final OCR token embedding $\{x^\text{ocr}_n\}$ as below
\begin{equation}
    x^\text{ocr}_n = \text{LN}(W_3 x^\text{ft}_n + W_4 x^\text{fr}_n + W_5 x^\text{p}_n) + \text{LN}(W_6 x^\text{b}_n)
\end{equation}
where $W_3$, $W_4$, $W_5$ and $W_6$ are learned projection matrices and $\text{LN}(\cdot)$ is layer normalization.

\subsection{Multimodal fusion and iterative answer prediction with pointer-augmented transformers}
\label{sec:method_ans_dec}

After embedding all entities (question words, visual objects, and OCR tokens) from each modality as vectors in the $d$-dimensional joint embedding space as described in Sec.~\ref{sec:method_embedding}, we apply a stack of $L$ transformer layers \cite{vaswani2017attention} with a hidden dimension of $d$ over the list of all $K+M+N$ entities from $\{x^\text{ques}_k\}$, $\{x^\text{obj}_m\}$, and $\{x^\text{ocr}_n\}$. Through the multi-head self-attention mechanism in transformers, each entity is allowed to freely attend to all other entities, regardless of whether they are from the same modality or not. For example, an OCR token is allowed to attend to another OCR token, a detected object, or a question word. This enables modeling both inter- and intra- modality relations in a homogeneous way through the same set of transformer parameters. The output from our multimodal transformer is a list of $d$-dimensional feature vectors for entities in each modality, which can be seen as their enriched embedding in multimodal context.

We predict an answer to the question through iterative decoding, using exactly the same transformer layers as a decoder. We decode the answer word by word in an auto-regressive manner for a total of $T$ steps, where each decoded word may be either an OCR token in the image or a word from our fixed vocabulary of frequent answer words. As illustrated in Figure~\ref{fig:method}, at each step during decoding, we feed in an embedding of the previously predicted word, and predict the next answer word based on the transformer output with a dynamic pointer network.

Let $\{z^\text{ocr}_1,\cdots,z^\text{ocr}_N\}$ be the $d$-dimensional transformer outputs of the $N$ OCR tokens in the image. Assume we have a vocabulary of $V$ words that frequently appear in the training set answers. At the $t$-th decoding step, the transformer model outputs a $d$-dimensional vector $z^\text{dec}_t$ corresponding to the input $x^\text{dec}_t$ at step $t$ (explained later in this section). From $z^\text{dec}_t$, we predict both the $V$-dimensional scores $y^\text{voc}_t$ of choosing a word from fixed answer vocabulary and the $N$-dimensional scores $y^\text{ocr}_t$ of selecting an OCR token from the image at decoding step $t$.
In our implementation, the fixed answer vocabulary score $y^\text{voc}_{t,i}$ for the $i$-th word (where $i=1,\cdots,V$) 
is predicted as a simple linear layer as
\begin{equation}
\label{eqn:scores_vocab}
y^\text{voc}_{t,i} = \left(w^\text{voc}_i\right)^T z^\text{dec}_t + b^\text{voc}_i
\end{equation}
where $w^\text{voc}_i$ is a $d$-dimensional parameter for the $i$-th word in the answer vocabulary, and $b^\text{voc}_i$ is a scalar parameter.

To select a token from the $N$ OCR tokens in the image, we augment the transformer model with a dynamic pointer network, predicting a copying score $y^\text{ocr}_{t,n}$ (where $n=1,\cdots,N$) for each token via bilinear interaction between the decoding output $z^\text{dec}_t$ and each OCR token's output representation $z^\text{ocr}_n$ as
\begin{equation}
\label{eqn:scores_ocr}
y^\text{ocr}_{t,n} = \left(W^\text{ocr} z^\text{ocr}_n + b^\text{ocr}\right)^T \left(W^\text{dec} z^\text{dec}_t + b^\text{dec}\right)
\end{equation}
where $W^\text{ocr}$ and $W^\text{dec}$ are $d\times d$ matrices, and $b^\text{ocr}$ and $b^\text{dec}$ are $d$-dimensional vectors.

During prediction, we take the argmax on the concatenation $y^\text{all}_t=[y^\text{voc}_t; y^\text{ocr}_t]$ of fixed answer vocabulary scores and dynamic OCR-copying scores, selecting the top scoring element (either a vocabulary word or an OCR token) from all $V+N$ candidates.

In our iterative auto-regressive decoding procedure, if the prediction at decoding time-step $t$ is an OCR token, we feed in its OCR representation $x^\text{ocr}_n$ as the transformer input $x^\text{dec}_{t+1}$ to the next prediction step $t+1$. Otherwise (the previous prediction is a word from the fixed answer vocabulary), we feed in its corresponding weight vector $w^\text{voc}_i$ in Eqn.~\ref{eqn:scores_vocab} as the next step's input $x^\text{dec}_{t+1}$. In addition, we add two extra $d$-dimensional vectors as inputs -- a positional embedding vector corresponding to step $t$, and a type embedding vector corresponding to whether the previous prediction is a fixed vocabulary word or an OCR token. Similar to machine translation, we augment our answer vocabulary with two special tokens, \texttt{<begin>} and \texttt{<end>}. Here \texttt{<begin>} is used as the input to the first decoding step, and we stop the decoding process after \texttt{<end>} is predicted.

To ensure causality in answer decoding, we mask the attention weights in the self-attention layers of the transformer architecture \cite{vaswani2017attention} such that question words, detected objects and OCR tokens cannot attend to any decoding steps, and all decoding steps can only attend to previous decoding steps in addition to question words, detected objects and OCR tokens. This is similar to prefix LM in \cite{raffel2019exploring}. 

\subsection{Training}
\label{sec:method_training}

During training, we supervise our multimodal transformer at each decoding step. Similar to sequence prediction tasks such as machine translation, we use teacher-forcing \cite{lamb2016professor} (\ie using ground-truth inputs to the decoder) to train our multi-step answer decoder, where each ground-truth answer is tokenized into a sequence of words. Given that an answer word can appear in both fixed answer vocabulary and OCR tokens, we apply multi-label sigmoid loss (instead of softmax loss) over the concatenated scores $y^\text{all}_t$.

\section{Experiments}
\label{sec:exp}

We evaluate our model on three challenging datasets for the TextVQA task, including TextVQA \cite{singh2019towards}, ST-VQA \cite{biten2019scene}, and OCR-VQA \cite{mishraocr19} (we use these datasets for research purposes only). Our model outperforms previous work by a significant margin on all the three datasets.

\subsection{Evaluation on the TextVQA dataset}
\label{sec:exp_textvqa}

\begin{table*}[t]
\vspace{\doublecoltopmargin}
\small
\begin{center}
\begin{tabular}{clcclccc}
\toprule
\multirow{2}{*}{\#} & \multirow{2}{*}{Method} & Question enc. & OCR & OCR token & Output & Accu. & Accu. \\
& & pretraining & system & representation & module & on val & on test \\
\midrule
1 & LoRRA \cite{singh2019towards} & GloVe & Rosetta-ml & FastText & classifier & 26.56 & 27.63 \\
\midrule
2 & \methodname w/o dec. & GloVe & Rosetta-ml & FastText & classifier & 29.36 & -- \\
3 & \methodname w/o dec. & (none) & Rosetta-ml & FastText & classifier & 29.55 & -- \\
4 & \methodname w/o dec. & BERT & Rosetta-ml & FastText & classifier & 30.15 & -- \\
5 & \methodname w/o dec. & BERT & Rosetta-en & FastText & classifier & 31.28 & -- \\
6 & \methodname w/o dec. & BERT & Rosetta-en & FastText + bbox & classifier & 33.32 & -- \\
7 & \methodname w/o dec. & BERT & Rosetta-en & FastText + bbox + FRCN & classifier & 34.38 & -- \\
8 & \methodname w/o dec. & BERT & Rosetta-en & FastText + bbox + FRCN + PHOC & classifier & \textbf{35.70} & -- \\
\midrule
9 & \methodname (ours - ablation) & (none) & Rosetta-ml & FastText + bbox + FRCN + PHOC & decoder & 36.06 & -- \\
10 & \methodname (ours - ablation) & BERT & Rosetta-ml & FastText + bbox + FRCN + PHOC & decoder & 37.06 & -- \\
11 & \methodname (ours) & BERT & Rosetta-en & FastText + bbox + FRCN + PHOC & decoder & \textbf{39.40} & 39.01 \\
\midrule
12 & DCD\_ZJU (ensemble) \cite{textvqa_dcd} & -- & -- & -- & -- & 31.48 & 31.44 \\
13 & MSFT\_VTI \cite{textvqa_msft_vti} & -- & -- & -- & -- & 32.92 & 32.46 \\
14 & \methodname (ours; w/ ST-VQA) & BERT & Rosetta-en & FastText + bbox + FRCN + PHOC & decoder & \textbf{40.55} & \textbf{40.46} \\
\bottomrule
\end{tabular}
\end{center}
\vspace{\tabcapmargin}
\caption{On the TextVQA dataset, we ablate our \methodname model and show a detailed comparison with prior work LoRRA \cite{singh2019towards}. Our multimodal transformer (line 3 vs 1), our rich OCR representation (line 8 vs 5) and our iterative answer prediction (line 11 vs 8) all improve the accuracy significantly. Notably, our model still outperforms LoRRA by 9.5\% (absolute) even when using fewer pretrained parameters (line 9 vs 1). Our final model achieves 39.01\% (line 11) and 40.46\% (line 14) test accuracy without and with the ST-VQA dataset as additional training data respectively, outperforming the challenge-winning DCD\_ZJU method by 9\% (absolute). See Sec.~\ref{sec:exp_textvqa} for details.}
\label{tab:exp_textvqa_detailed_comparison}
\vspace{\tabmargin}
\end{table*}

The TextVQA dataset \cite{singh2019towards} contains 28,408 images from the Open Images dataset \cite{kuznetsova2018open}, with human-written questions asking to reason about text in the image. Similar to VQAv2 \cite{goyal2017making}, each question in the TextVQA dataset has 10 human annotated answers, and the final accuracy is measured via soft voting of the 10 answers.\footnote{See \url{https://visualqa.org/evaluation} for details.}

We use $d=768$ as the dimensionality of the joint embedding space and extract question word features with BERT-BASE using the 768-dimensional outputs from its first three layers, which are fine-tuned during training.

For visual objects, following Pythia \cite{singh2018pythia} and LoRRA \cite{singh2019towards}, we detect objects with a Faster R-CNN detector \cite{ren2015faster} pretrained on the Visual Genome dataset \cite{krishna2017visual}, and keeps 100 top-scoring objects per image. Then, the fc6 feature vector is extracted from each detected object. We apply the Faster R-CNN fc7 weights on the extracted fc6 features to output 2048-dimensional fc7 appearance features and fine-tune fc7 weights during training. However, we do not use the ResNet-152 convolutional features \cite{he2016deep} as in LoRRA.

Finally, we extract text tokens on each image using the Rosetta OCR system \cite{borisyuk2018rosetta}. Unlike the prior work LoRRA \cite{singh2019towards} that uses a multilingual Rosetta version, in our model we use an English-only version of Rosetta that we find has higher recall. 
We refer to these two versions as \textbf{Rosetta-ml} and \textbf{Rosetta-en}, respectively. As mentioned in Sec.~\ref{sec:method_embedding}, from each OCR token we extract \textbf{FastText} \cite{bojanowski2017enriching} feature, appearance feature from Faster R-CNN (\textbf{FRCN}), \textbf{\textbf{PHOC}} \cite{almazan2014word} feature, and bounding box (\textbf{bbox}) feature.

In our multimodal transformer, we use $L=4$ layers of multimodal transformer with 12 attention heads. Other hyper-parameters (such as dropout ratio) follow BERT-BASE \cite{devlin2019bert}. However, we note that the multimodal transformer parameters are initialized from scratch rather than from a pretrained BERT model. We use $T=12$ maximum decoding step in answer prediction unless otherwise specified, which is sufficient to cover almost all answers.

We collect the top 5000 frequent words from the answers in the training set as our answer vocabulary. During training, we use a batch size of 128, and train for a maximum of 24,000 iterations. Our model is trained using the Adam optimizer, with a learning rate of 1e-4 and a staircase learning rate schedule, where we multiply the learning rate by 0.1 at 14000 and at 19000 iterations. The best snapshot is selected using the validation set accuracy. The entire training takes approximately 10 hours on 4 Nvidia Tesla V100 GPUs.

As a notable prior work on this dataset, we show a step-by-step comparison with the LoRRA model \cite{singh2019towards}. LoRRA uses two single-hop attention layers over image visual features and OCR features. The attended visual and OCR features are then fused with a vector encoding of the question and fed into a single-step classifier to select either a frequent answer from the training set or a single OCR token from the image. Unlike our rich OCR representation in Sec.~\ref{sec:method_embedding}, in the LoRRA model each OCR token is only represented as a 300-dimensional FastText vector.

\myparagraph{Ablations on pretrained question encoding and OCR systems} We first experiment with a restricted version of our model using the multimodal transformer architecture but without iterative decoding in answer prediction, \ie \textbf{\methodname (w/o dec.)} in Table~\ref{tab:exp_textvqa_detailed_comparison}. In this setting, we only decode for one step, and either select a frequent answer\footnote{In this case, we predict the entire (multi-word) answer, instead of a single word from our answer word vocabulary as in our full model.} from the training set or copy a single OCR token in the image as the answer. As a step-by-step comparison with LoRRA, we start with extracting OCR tokens from Rosetta-ml, representing OCR tokens only with FastText vectors, and initializing question encoding parameters in Sec.~\ref{sec:method_embedding} from scratch (rather than from a pretrained BERT-BASE model). The result is shown in line 3 of Table~\ref{tab:exp_textvqa_detailed_comparison}. Compared with LoRRA in line 1, this restricted version of our model already outperforms LoRRA by around 3\% (absolute) on TextVQA validation set. This result shows that our multimodal transformer architecture is more efficient for jointly modeling the three input modalities. We also experiment with initializing the word embedding from GloVe \cite{pennington2014glove} as in LoRRA and the remaining parameters from scratch, shown in line 2. However, we find that this setting slightly under-performs initializing everything from scratch, which we suspect is due to different question tokenization between LoRRA and the BERT tokenizer used in our model. We then switch to a pretrained BERT for question encoding in line 4, and Rosetta-en for OCR extraction in line 5. Comparing line 3 to 5, we see that a pretrained BERT leads to around 0.6\% higher accuracy, and Rosetta-en gives another 1\% improvement.

\begin{figure}[t]
\centering
\includegraphics[width=0.9\linewidth,trim=0 5 0 5,clip=false]{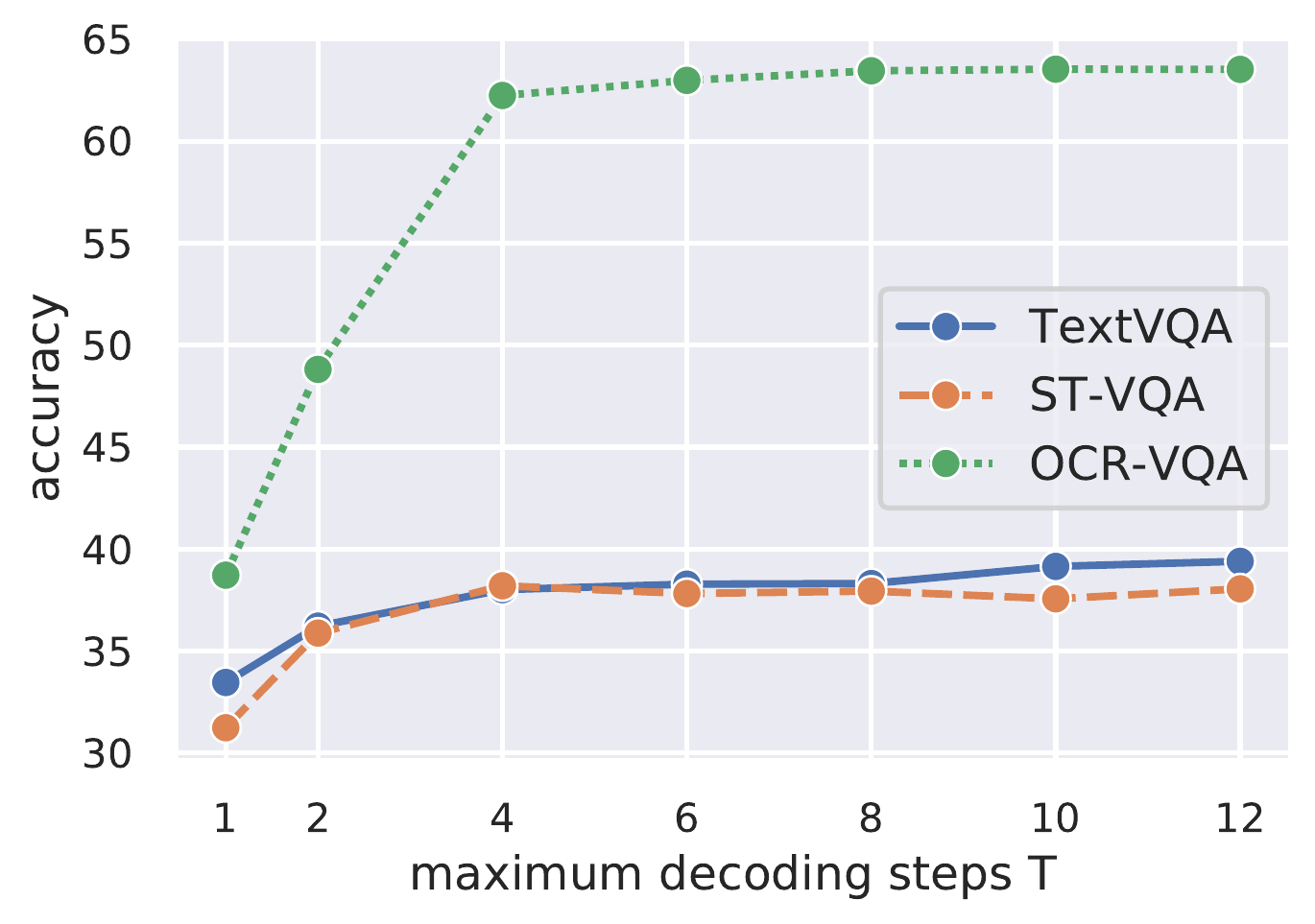}
\caption{Accuracy under different maximum decoding steps $T$ on the validation set of TextVQA, ST-VQA, and OCR-VQA. There is a major gap between single-step ($T=1$) and multi-step ($T>1$) answer prediction. We use 12 steps by default in our experiments.}
\label{fig:exp_dec_step_num}
\vspace{\figmargin}
\end{figure}

\paragraph{Ablations on OCR feature representation} We analyze the impact of our rich OCR representation in Sec.~\ref{sec:method_embedding} through ablations in Table~\ref{tab:exp_textvqa_detailed_comparison} line 5 to 8. We see that OCR location (bbox) features and the RoI-pooled appearance features (FRCN) both improve the performance by a noticeable margin. In addition, we find that PHOC is also helpful as a character-level representation of the OCR token. Our rich OCR representation gives around 4\% (absolute) accuracy improvement compare with using only FastText features as in LoRRA (line 8 vs 5). We note that our extra OCR features do not require more pretrained models, as we apply exactly the same Faster R-CNN model use in object detection for OCR appearance features, and PHOC is a manually-designed feature that does not need pretraining.

\begin{figure*}[t]
\vspace{\doublecoltopmargin}
\footnotesize
\centering
\begin{tabular}{p{0.25\linewidth}p{0.22\linewidth}p{0.22\linewidth}p{0.23\linewidth}}
\includegraphics[width=\linewidth]{00000012_fa9ffd5aca1e4e51_34614_matched_multi_im.jpg} &
\includegraphics[width=\linewidth]{00000082_9742c58a933a5a1e_34684_matched_multi_im.jpg} &
\includegraphics[width=\linewidth]{00000125_9bf5082c0219d773_34727_matched_multi_im.jpg} &
\includegraphics[width=\linewidth]{00000006_181f00d3ee2b2076_34608_unmatched_single_im.jpg} \\
\textit{What does the light sign read on the farthest right window?} &
\textit{Who is usa today's bestselling author?} &
\textit{What is the name of the band?} &
\textit{what is the time?} \\
LoRRA: \vocab{exit} &
LoRRA: \vocab{roger zelazny} & 
LoRRA: \vocab{7} &
LoRRA: \vocab{1:45} \\
\methodname (ours): \ocr{bud} \vocab{light} &
\methodname (ours): \ocr{cathy} \ocr{williams} & 
\methodname (ours): \ocr{soul} \ocr{doubt} &
\methodname (ours): \vocab{3:44} \\
human: \textbf{bud light}; \textbf{all 2 liters}&
human: \textbf{cathy williams} &
human: \textbf{soul doubt}; \textbf{h. michael karshis}; \textbf{unanswerable} &
human: \textbf{5:40}; \textbf{5:41}; \textbf{5:42}; \textbf{8:00} \\
\end{tabular}
\vspace{\figcapmargin}
\caption{Qualitative examples from our \methodname model on the TextVQA validation set (\ocr{orange} words are from OCR tokens and \vocab{blue} words are from fixed answer vocabulary). Compared to the previous work LoRRA \cite{singh2019towards} which selects one answer from training set or copies only a single OCR token, our model can copy multiple OCR tokens and combine them with its fixed vocabulary through iterative decoding.}
\label{fig:vis_textvqa}
\vspace{\figmargin}
\end{figure*}

\myparagraph{Iterative answer decoding} We then apply our full \methodname model with iterative answer decoding to the TextVQA dataset. The results are shown in Table~\ref{tab:exp_textvqa_detailed_comparison} line 11, which is around 4\% (absolute) higher than its counterpart in line 8 using a single-step classifier and 13\% (absolute) higher than LoRRA in line 1. In addition, we ablate our model using Rosetta-ml and randomly initialized question encoding parameters in line 9 and 10. Here, we see that our model in line 9 still outperforms LoRRA (line 1) by as much as 9.5\% (absolute) when using the same OCR system as LoRRA and even fewer pretrained components. We also analyze the performance of our model with respect to the maximum decoding steps, shown in Figure~\ref{fig:exp_dec_step_num}, where decoding for multiple steps greatly improves the performance compared with a single step. Figure~\ref{fig:vis_textvqa} shows qualitative examples (more examples in appendix) of our \methodname model on the TextVQA dataset in comparison to LoRRA \cite{singh2019towards}, where our model is capable of selecting multiple OCR tokens and combining them with its fixed vocabulary in predicted answers.

\myparagraph{Qualitative insights} When inspecting the errors, we find that a major source of errors is OCR failure (\eg in the last example in Figure~\ref{fig:vis_textvqa}, we find that the digits on the watch are not detected). This suggests that the accuracy of our model could be improved with better OCR systems, as supported by the comparison between line 10 and 11 in Table~\ref{tab:exp_textvqa_detailed_comparison}. Another possible future direction is to dynamically recognize text in the image based on the question (\eg if the question asks about the price of a product brand, one may want to directly localize the brand name in the image). Some other errors of our model include resolving relations between objects and text or understanding large chunks of text in images (such as book pages). However, our model is able to correct a large number of mistakes in previous work where copying multiple text tokens is required to form an answer.

\myparagraph{TextVQA Challenge 2019} We also compare to the winning entries in the TextVQA Challenge 2019.\footnote{\url{https://textvqa.org/challenge}} We compare our method to DCD \cite{textvqa_dcd} (the challenge winner, based on ensemble) and MSFT\_VTI \cite{textvqa_msft_vti} (the top entry after the challenge), both relying on one-step prediction. We show that our single model (line 11) significantly outperforms these challenge winning entries on the TextVQA test set by a large margin. We also experiment with using the ST-VQA dataset \cite{biten2019scene} as additional training data (a practice used by some of the previous challenge participants), which gives another 1\% improvement and 40.46\% final test accuracy -- a new state-of-the-art on the TextVQA dataset.

\subsection{Evaluation on the ST-VQA dataset}
\label{sec:exp_stvqa}

\begin{figure*}[t]
\vspace{\doublecoltopmargin}
\footnotesize
\centering
\begin{tabular}{p{0.25\linewidth}p{0.187\linewidth}p{0.25\linewidth}p{0.235\linewidth}}
\includegraphics[width=\linewidth]{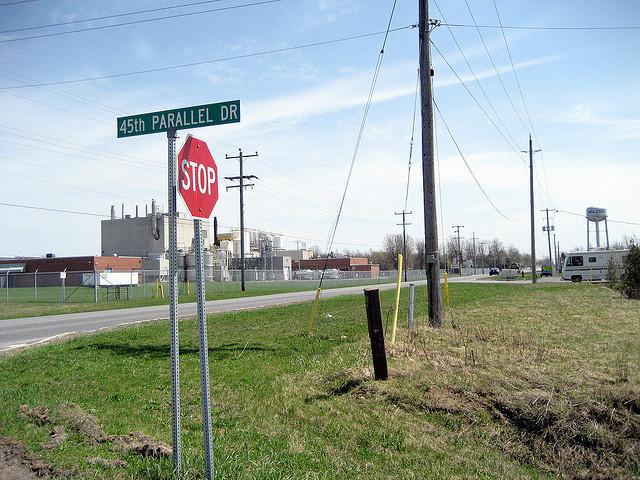} &
\includegraphics[width=\linewidth]{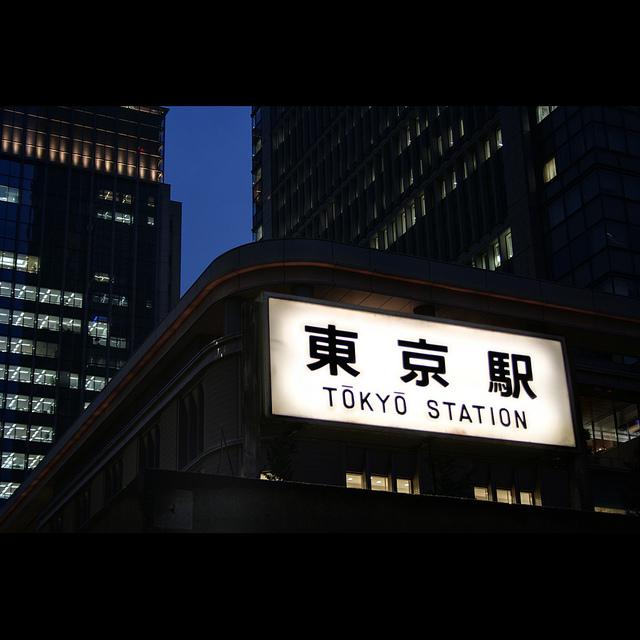} &
\includegraphics[width=\linewidth]{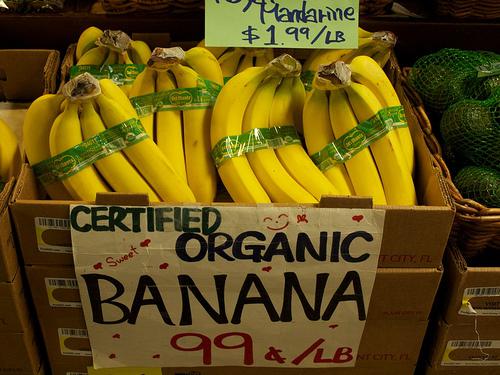} &
\includegraphics[width=\linewidth]{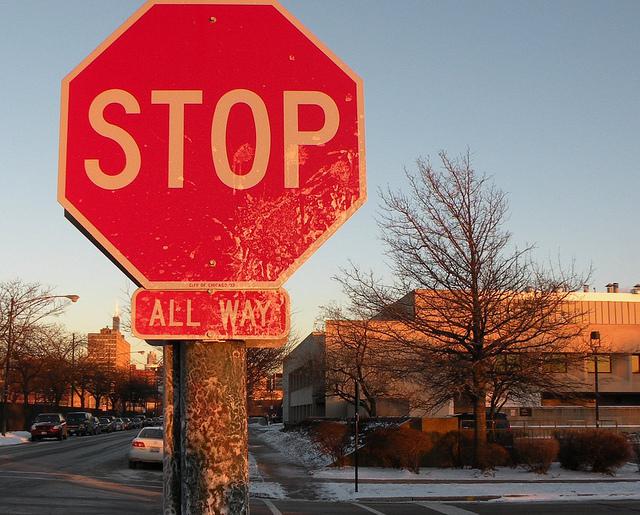} \\
\textit{What is the name of the street on which the Stop sign appears?} &
\textit{What does the white sign say?} &
\textit{How many cents per pound are the bananas?} &
\textit{What kind of stop sign is in the image?} \\
prediction: \ocr{45th} \ocr{parallel} \vocab{dr} &
prediction: \ocr{tokyo} \vocab{station} & 
prediction: \vocab{99} &
prediction: \vocab{stop} \vocab{all} \ocr{way} \\
GT: \textbf{45th parallel dr} &
GT: \textbf{tokyo station} &
GT: \textbf{99} &
GT: \textbf{all way} \\
\end{tabular}
\vspace{\figcapmargin}
\caption{Qualitative examples from our \methodname model on the ST-VQA validation set (\ocr{orange} words from OCR tokens and \vocab{blue} words from fixed answer vocabulary). Our model can select multiple OCR tokens and combine them with its fixed vocabulary to predict an answer.}
\label{fig:vis_stvqa}
\vspace{\figmargin}
\end{figure*}

The ST-VQA dataset \cite{biten2019scene} contains natural images from multiple sources including ICDAR 2013 \cite{karatzas2013icdar}, ICDAR 2015 \cite{karatzas2015icdar}, ImageNet \cite{deng2009imagenet}, VizWiz \cite{gurari2018vizwiz}, IIIT STR \cite{mishra2013image}, Visual Genome \cite{krishna2017visual}, and COCO-Text \cite{veit2016coco}.\footnote{We notice that many images from COCO-Text \cite{veit2016coco} in the downloaded ST-VQA data (around 1/3 of all images) are resized to $256 \times 256$ for unknown reasons, which degrades the image quality and distorts their aspect ratios. In our experiments, we replace these images with their original versions from COCO-Text as inputs to object detection and OCR systems.} The format of the ST-VQA dataset is similar to the TextVQA dataset in Sec.~\ref{sec:exp_textvqa}. However, each question is accompanied by only one or two ground-truth answers provided by the question writer. The dataset involves three tasks, and its Task 3 - Open Dictionary (containing 18,921 training-validation images and test 2,971 images) corresponds to our general TextVQA setting where no answer candidates are provided at test time.

The ST-VQA dataset adopts Average Normalized Levenshtein Similarity (ANLS)\footnote{\url{https://rrc.cvc.uab.es/?ch=11&com=tasks}\label{fn:anls}} as its official evaluation metric, defined as scores $1 - d_L(a_\text{pred}, a_\text{gt}) / \max(|a_\text{pred}|, |a_\text{gt}|)$ (where $a_\text{pred}$ and $a_\text{gt}$ are prediction and ground-truth answers and $d_L$ is edit distance) averaged over all questions. Also, all scores below the threshold 0.5 are truncated to 0 before averaging. To facilitate comparison, we report both accuracy and ANLS in our experiments.

As the ST-VQA dataset does not have an official split for training and validation, we randomly select 17,028 images as our training set and use the remaining 1,893 images as our validation set. We train our model on the ST-VQA dataset following exactly the same setting (line 11 in Table~\ref{tab:exp_textvqa_detailed_comparison}) as in our TextVQA experiments in Sec.~\ref{sec:exp_textvqa}, where we extract image text tokens using Rosetta-en, use FastText + bbox + FRCN + PHOC as our OCR representation, and initialize question encoding parameters from a pretrained BERT-BASE model. The results are shown in Table~\ref{tab:exp_stvqa_detailed_comparison}.

\begin{table}[t]
\small
\begin{center}
\begin{tabular}{clcccc}
\toprule
\multirow{2}{*}{\#} & \multirow{2}{*}{Method} & Output & Accu. & ANLS & ANLS \\
& & module & on val & on val & on test \\
\midrule
1 & SAN+STR \cite{biten2019scene} & -- & -- & -- & 0.135 \\
2 & VTA \cite{biten2019icdar} & -- & -- & -- & 0.282 \\
\midrule
3 & \methodname w/o dec. & classifier & 33.52 & 0.397 & -- \\
4 & \methodname (ours) & decoder & \textbf{38.05} & \textbf{0.472} & \textbf{0.462} \\
\bottomrule
\end{tabular}
\end{center}
\vspace{\tabcapmargin}
\caption{On the ST-VQA dataset, our restricted model without decoder (\methodname w/o dec.) already outperforms previous work by a large margin. Our final model achieves +0.18 (absolute) ANLS boost over the challenge winner, VTA \cite{biten2019icdar}. See Sec.~\ref{sec:exp_stvqa} for details.}
\label{tab:exp_stvqa_detailed_comparison}
\vspace{\tabmargin}
\end{table}

\myparagraph{Ablations of our model} We train two versions of our model, one restricted version (\methodname w/o dec. in Table~\ref{tab:exp_stvqa_detailed_comparison}) with a fixed one-step classifier as output module (similar to line 8 in Table~\ref{tab:exp_textvqa_detailed_comparison}) and one full version (\methodname) with iterative answer decoding. Comparing the results of these two models, it can be seen that there is a large improvement from our iterative answer prediction mechanism.

\myparagraph{Comparison to previous work} We compare with two previous methods on this dataset: 1) SAN+STR \cite{biten2019scene}, which combines SAN for VQA \cite{yang2016stacked} and Scene Text Retrieval \cite{gomez2018single} for answer vocabulary retrieval, and 2) VTA \cite{biten2019icdar}, the ICDAR 2019 ST-VQA Challenge\footref{fn:anls} winner, based on BERT  \cite{devlin2019bert} for question encoding and BUTD \cite{anderson2018bottom} for VQA. From Table~\ref{tab:exp_stvqa_detailed_comparison}, it can be seen that our restricted model (\methodname w/o dec.) already achieves higher ANLS than these two models, and our full model achieves as much as +0.18 (absolute) ANLS boost over the best previous work.

We also ablate the maximum copying number in our model in Figure~\ref{fig:exp_dec_step_num}, showing that it is beneficial to decode for multiple (as opposed to one) steps. Figure~\ref{fig:vis_stvqa} shows qualitative examples of our model on the ST-VQA dataset.

\subsection{Evaluation on the OCR-VQA dataset}
\label{sec:exp_ocrvqa}

\begin{table}[t]
\small
\begin{center}
\begin{tabular}{clccc}
\toprule
\multirow{2}{*}{\#} & \multirow{2}{*}{Method} & Output & Accu. & Accu. \\
& & module & on val & on test \\
\midrule
1 & BLOCK \cite{mishraocr19} & -- & -- & 42.0 \\
2 & CNN \cite{mishraocr19} & -- & -- & 14.3 \\
3 & BLOCK+CNN \cite{mishraocr19} & -- & -- & 41.5 \\
4 & BLOCK+CNN+W2V \cite{mishraocr19} & -- & -- & 48.3 \\
\midrule
5 & \methodname w/o dec. & classifier & 46.3 & -- \\
6 & \methodname (ours) & decoder & \textbf{63.5} & \textbf{63.9} \\
\bottomrule
\end{tabular}
\end{center}
\vspace{\tabcapmargin}
\caption{On the OCR-VQA dataset, we experiment with using either an iterative decoder (our full model) or a single-step classifier (\methodname w/o dec.) as the output module, where our iterative decoder greatly improves the accuracy and largely outperforms the baseline methods. See Sec.~\ref{sec:exp_ocrvqa} for details.}
\label{tab:exp_ocrvqa_detailed_comparison}
\end{table}

\begin{figure}[t]
\footnotesize
\centering
\begin{tabular}{p{0.5\linewidth}p{0.423\linewidth}}
\includegraphics[width=\linewidth]{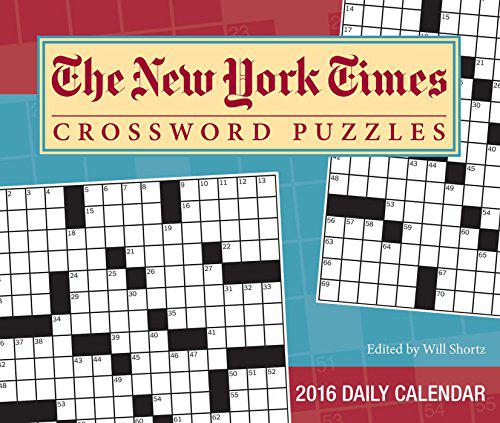} &
\includegraphics[width=\linewidth]{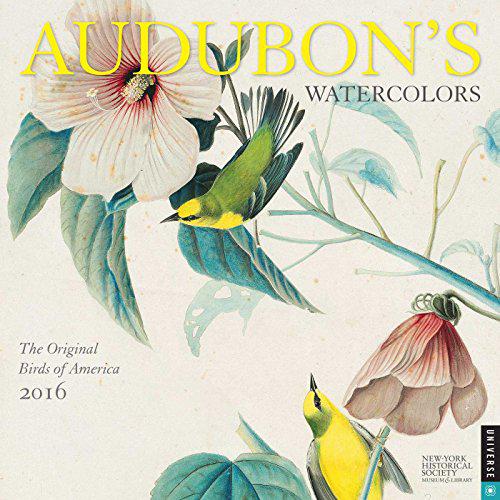} \\
\textit{Who is the author of this book?} &
\textit{Is this a pharmaceutical book?} \\
prediction: \ocr{the} \ocr{new} \ocr{york} \ocr{times} &
prediction: \vocab{no} \\
GT: \textbf{the new york times} &
GT: \textbf{no} \\
\end{tabular}
\vspace{\figcapmargin}
\caption{Qualitative examples from our \methodname model on the OCR-VQA validation set (\ocr{orange} words from OCR tokens and \vocab{blue} words from fixed answer vocabulary).}
\label{fig:vis_ocrvqa}
\vspace{\figmargin}
\end{figure}

The OCR-VQA dataset \cite{mishraocr19} contains 207,572 images of book covers, with template-based questions asking about the title, author, edition, genre, year or other information about the book. Each question is has a single ground-truth answer, and the dataset assumes that the answers to these questions can be inferred from the book cover images.

We train our model using the same hyper-parameters as in Sec.~\ref{sec:exp_textvqa} and \ref{sec:exp_stvqa}, but use $2\times$ the total iterations and adapted learning rate schedule since the OCR-VQA dataset contains more images. The results are shown in Table~\ref{tab:exp_ocrvqa_detailed_comparison}. Compared to using a one-step classifier (\methodname w/o dec.), our full model with iterative decoding achieves significantly better accuracy, which coincides with Figure~\ref{fig:exp_dec_step_num} that having multiple decoding steps is greatly beneficial on this dataset. This is likely because the OCR-VQA dataset often contains multi-word answers such as book titles and author names.

We compare to four baseline approaches from \cite{mishraocr19}, which are VQA systems based on 1) visual features from a convolutional network (CNN), 2) grouping OCR tokens into text blocks (BLOCK) with manually defined rules, 3) an averaged word2vec (W2V) feature over all the OCR tokens in the image, and 4) their combinations. Note that while the BLOCK baseline can also select multiple OCR tokens, it relies on manually defined rules to merge tokens into groups and can only select one group as answer, while our method learns from data how to copy OCR tokens to compose answers. Compare to these baselines, our \methodname has over 15\% (absolute) higher test accuracy. Figure~\ref{fig:vis_ocrvqa} shows qualitative examples of our model on this dataset.

\section{Conclusion}
\label{sec:conclusion}
In this paper, we present \methodfullname (\methodname) for visual question answering based on understanding and reasoning about text in images. \methodname adopts rich representations for text in the images, jointly models all modalities through a pointer-augmented multimodal transformer architecture over a joint embedding space, and predicts the answer through iterative decoding, outperforming previous work by a large margin on three challenging datasets for the TextVQA task.
Our results suggest that it is efficient to handle multiple modalities through domain-specific embedding followed by homogeneous self-attention and to generate complex answers as multi-step decoding instead of one-step classification.

{\small
\bibliographystyle{ieee_fullname}
\bibliography{textvqa}
}

\appendix
\counterwithin{figure}{section}
\counterwithin{table}{section}
\counterwithin{equation}{section}

\twocolumn[{
\begin{center}
\Large 
\textbf{Iterative Answer Prediction with Pointer-Augmented\\Multimodal Transformers for TextVQA}\\(Supplementary Material)
\par
\end{center}
\vspace{2em}
}]

\section{Hyper-parameters in \methodname}

We summarize the hyper-parameters in our \methodname model in Table~\ref{tab:supp_hyperparameters}. Most hyper-parameters are the same across all the three datasets (TextVQA, ST-VQA, and OCR-VQA), except that we use $2\times$ the total iterations and adapted learning rate schedule on the OCR-VQA dataset since it contains more images.

\begin{table}[h]
\begin{center}
\begin{tabular}{lc}
\toprule
Hyper-parameter & Value \\
\midrule
max question word num $K$ & 20 \\
detected object num $M$ & 100 \\
max OCR num $N$ & 50 \\
max decoding steps $T$ & 12 \\
embedding dim $d$ & 768 \\
multimodal transformer layers $L$ & 4 \\
multimodal transformer attention heads & 12 \\
multimodal transformer FFN dim & 3072 \\
multimodal transformer dropout & 0.1 \\
optimizer & Adam \\
batch size & 128 \\
base learning rate & 1e-4 \\
warm-up learning rate factor & 0.2 \\
warm-up iterations & 2000 \\
max gradient L2-norm for clipping & 0.25 \\
learning rate decay & 0.1 \\
learning rate steps (TextVQA, ST-VQA) & 14000, 19000 \\
learning rate steps (OCR-VQA) & 28000, 38000 \\
max iterations (TextVQA, ST-VQA) & 24000 \\
max iterations (OCR-VQA) & 48000 \\
\bottomrule
\end{tabular}
\end{center}
\caption{Hyper-parameters of our \methodname.}
\label{tab:supp_hyperparameters}
\end{table}

\section{Additional ablation analysis}

During the iterative answer decoding process, at each step our \methodname model can decode an answer word either from the model's fixed vocabulary, or from the OCR tokens extracted from the image. We find in our experiments that it is necessary to have \textit{both} the fixed vocabulary space and the OCR tokens.

Table~\ref{tab:supp_ablation_rmvocab_rmocr} shows our ablation study where we remove the fixed answer vocabulary or the dynamic pointer network for OCR copying from our \methodname. Both these two ablated versions have a large accuracy drop compared to our full model. However, we note that even without fixed answer vocabulary, our restricted model (\textbf{\methodname w/o fixed vocabulary} in Table~\ref{tab:supp_ablation_rmvocab_rmocr}) still outperforms the previous work LoRRA \cite{singh2019towards}, suggesting that it is particularly important to learn to copy multiple OCR tokens to form an answer (a key feature in our model but not in LoRRA).

\begin{table}[h]
\begin{center}
\begin{tabular}{clc}
\toprule
\# & Method & TextVQA Val Accuracy \\
\midrule
1 & LoRRA \cite{singh2019towards} & 26.56 \\
\midrule
2 & \methodname w/o fixed vocabulary & 31.76 \\ 
3 & \methodname w/o OCR copying & 14.94 \\
4 & \methodname (ours) & \textbf{39.40} \\
\bottomrule
\end{tabular}
\end{center}
\caption{We ablate our \methodname model by removing its fixed answer vocabulary (\methodname w/o fixed vocabulary) or its dynamic pointer network for OCR copying (\methodname w/o OCR copying) on the TextVQA dataset. We see that our full model has significantly higher accuracy than these ablations, showing that it is important to have \textit{both} a fixed and a dynamic vocabulary (\ie OCR tokens).}
\label{tab:supp_ablation_rmvocab_rmocr}
\end{table}

\section{Additional qualitative examples}

\begin{figure*}[t]
\footnotesize
\centering
\begin{tabular}{p{0.4\linewidth}p{0.4\linewidth}}
\includegraphics[width=\linewidth]{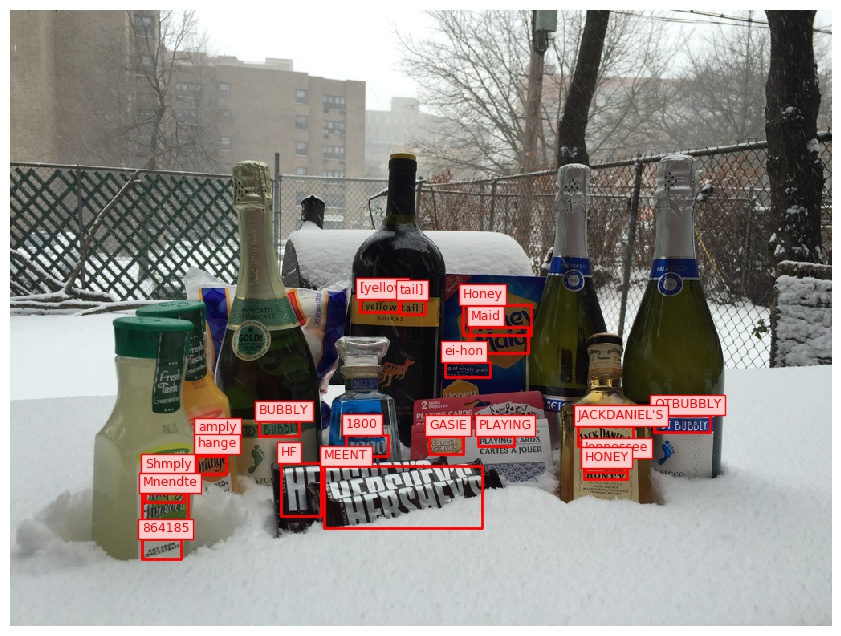} &
\includegraphics[width=\linewidth]{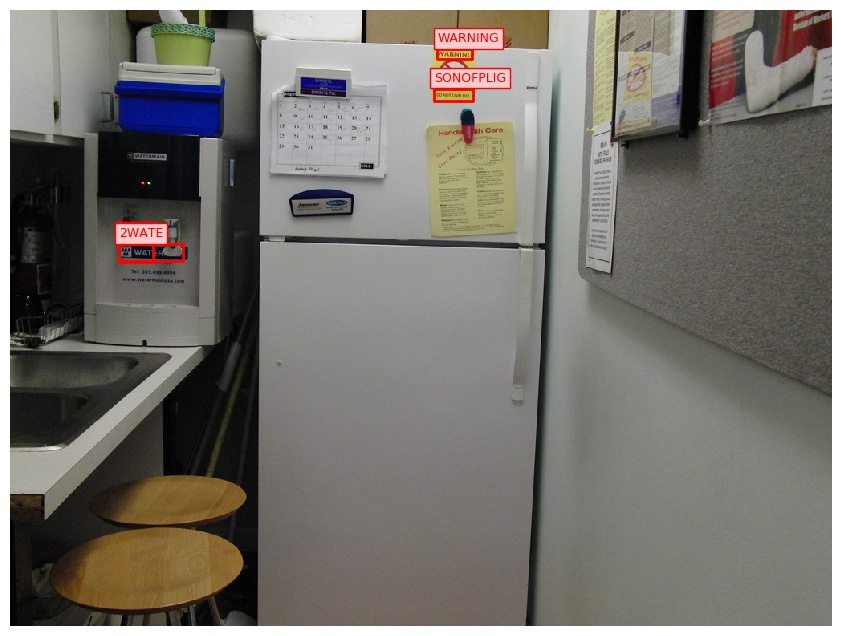} \\
(a) \textit{what candy bar is down there on the bottom?} &
(b) \textit{what is the year on the calender?} \\
prediction: \vocab{unanswerable} &
prediction: \vocab{2005} \\
human: \textbf{hershey's}; \textbf{hersheys} &
human: \textbf{2010}; \textbf{unanswerable} \\
\\

\includegraphics[width=\linewidth]{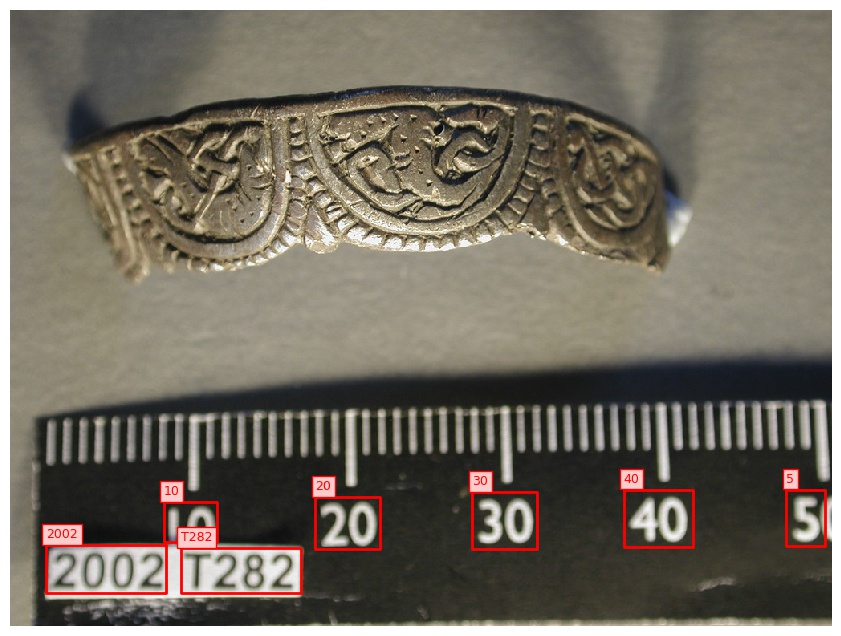} &
\includegraphics[width=\linewidth]{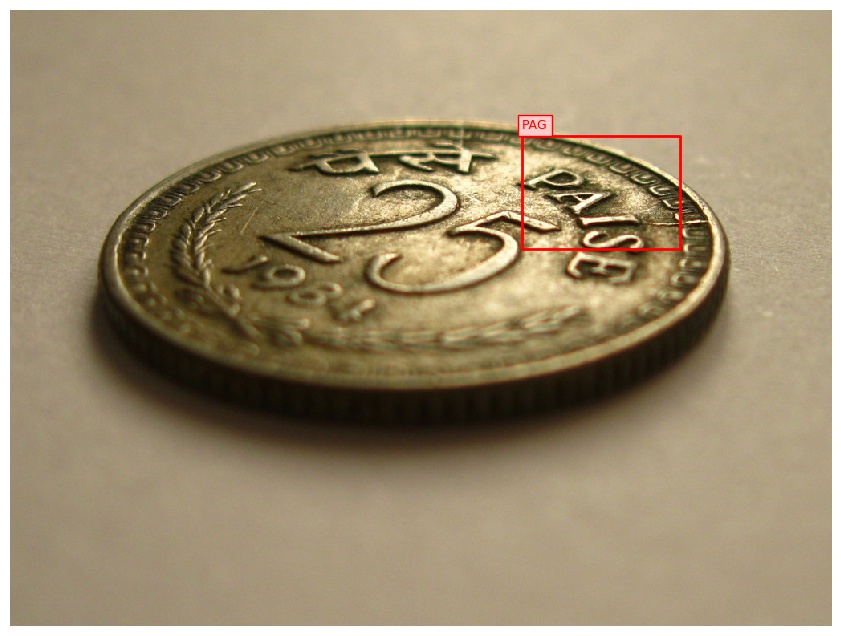} \\
(c) \textit{what is the largest measurement we can see on this ruler?} &
(d) \textit{how much is the coin worth?} \\
prediction: \ocr{40} &
prediction: \vocab{one} \vocab{dollar} \\
human: \textbf{50} &
human: \textbf{20}; \textbf{25}; \textbf{25 paise} \\
\\

\includegraphics[width=\linewidth]{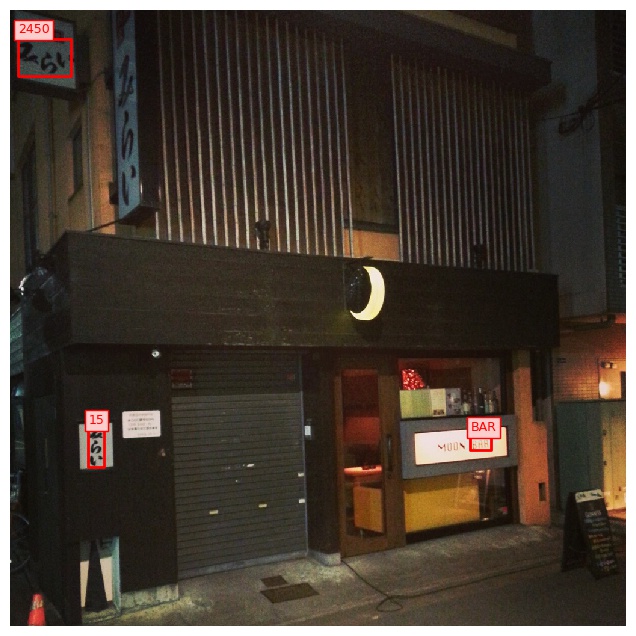} &
\includegraphics[width=\linewidth]{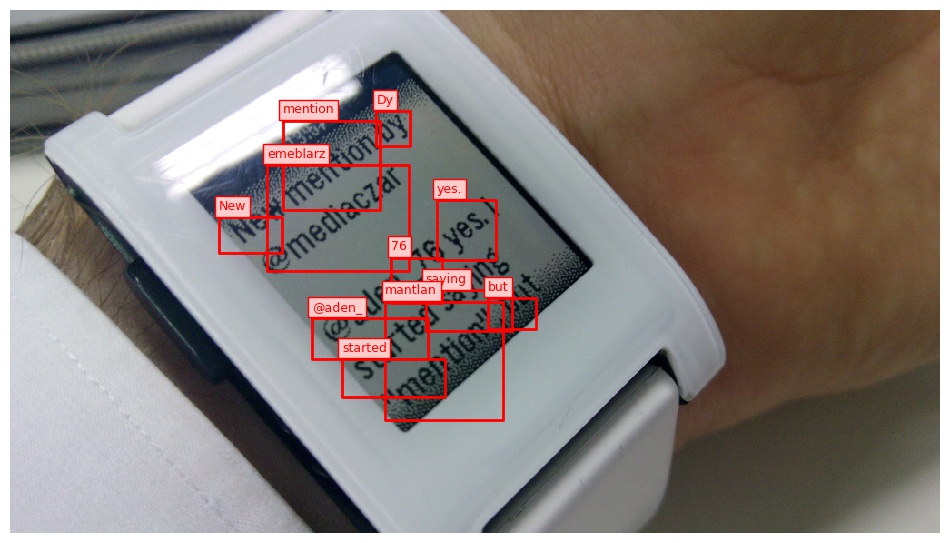} \\
(e) \textit{what is the name of the bar?} &
(f) \textit{what time is it?} \\
prediction: \vocab{15} &
prediction: \ocr{76} \\
human: \textbf{moo bar}; \textbf{moon}; \textbf{moon bar} &
human: \textbf{13:50}; \textbf{13:57}; \textbf{;5713}; \textbf{mathematic}; \textbf{wifi} \\
\end{tabular}
\vspace{0.5em}
\caption{Examples where OCR failure is the main source of errors (from our \methodname model on the TextVQA validation set). The red boxes show the OCR results (\ocr{orange} words from OCR tokens and \vocab{blue} words from fixed answer vocabulary).}
\label{fig:vis_supp_textvqa_ocrfail}
\end{figure*}

\begin{figure*}[t]
\vspace{\doublecoltopmargin}
\footnotesize
\centering
\begin{tabular}{p{0.22\linewidth}p{0.22\linewidth}p{0.22\linewidth}p{0.22\linewidth}}
\includegraphics[width=\linewidth]{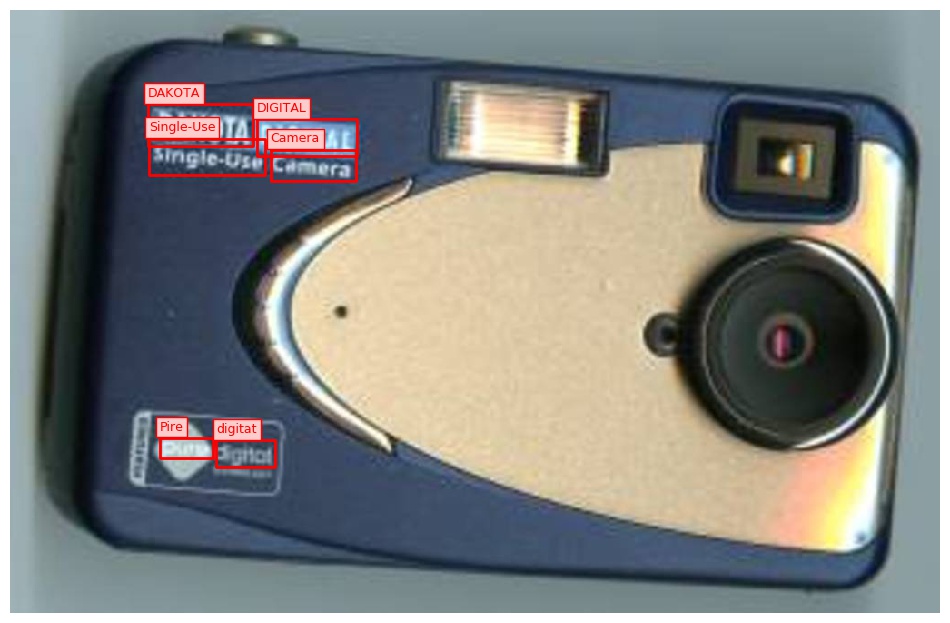} &
\includegraphics[width=\linewidth]{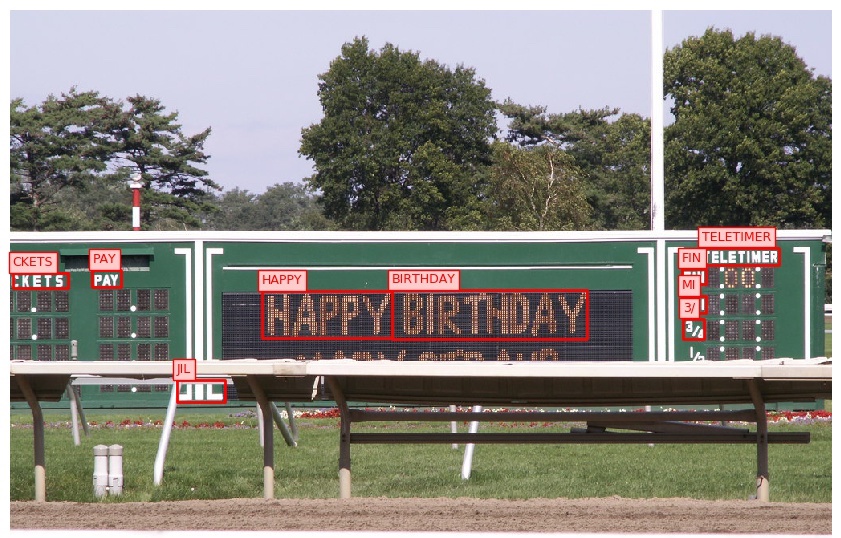} &
\includegraphics[width=\linewidth]{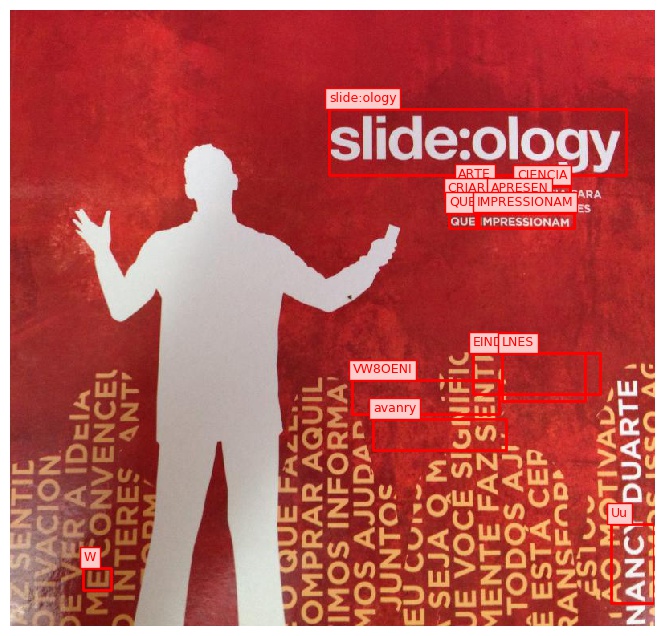} &
\includegraphics[width=\linewidth]{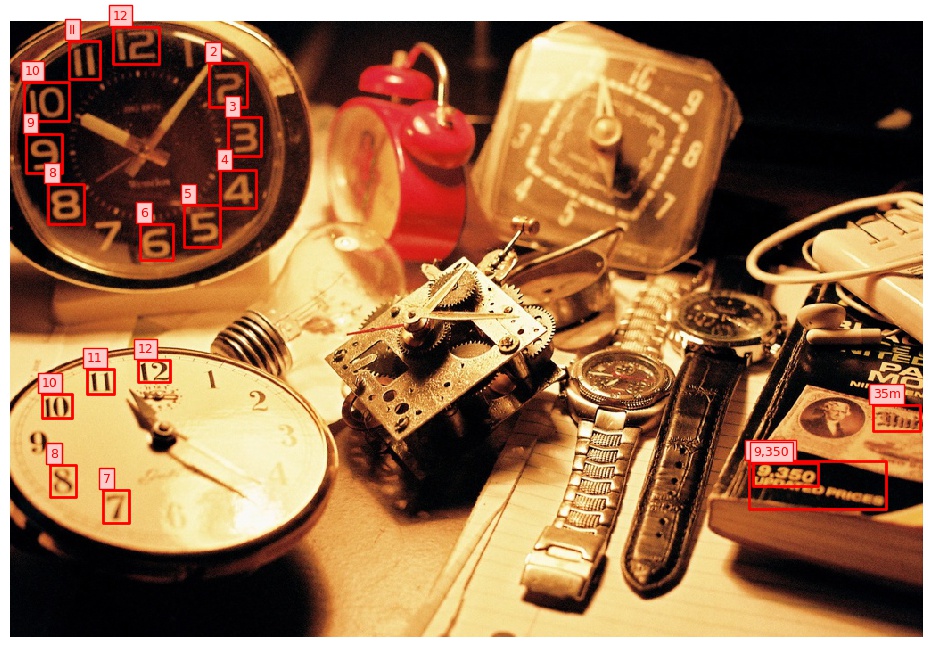} \\
(a) \textit{what is the brand of this camera?} &
(b) \textit{does it say happy birthday?} &
(c) \textit{what is the title of the album?} &
(d) \textit{what is the 4 digit number written at the bottom of the black book?} \\
\methodname: \ocr{dakota} \vocab{digital} &
\methodname: \vocab{yes} &
\methodname: \ocr{slide:ology} &
\methodname: \ocr{9350} \\
human: \textbf{dakota digital}; \textbf{dakota}; \textbf{clos culombu}; \textbf{nous les gosses} &
human: \textbf{yes} &
human: \textbf{slide:ology}; \textbf{sideology} &
human: \textbf{9350}; \textbf{9,350} \\
\\
\includegraphics[width=\linewidth]{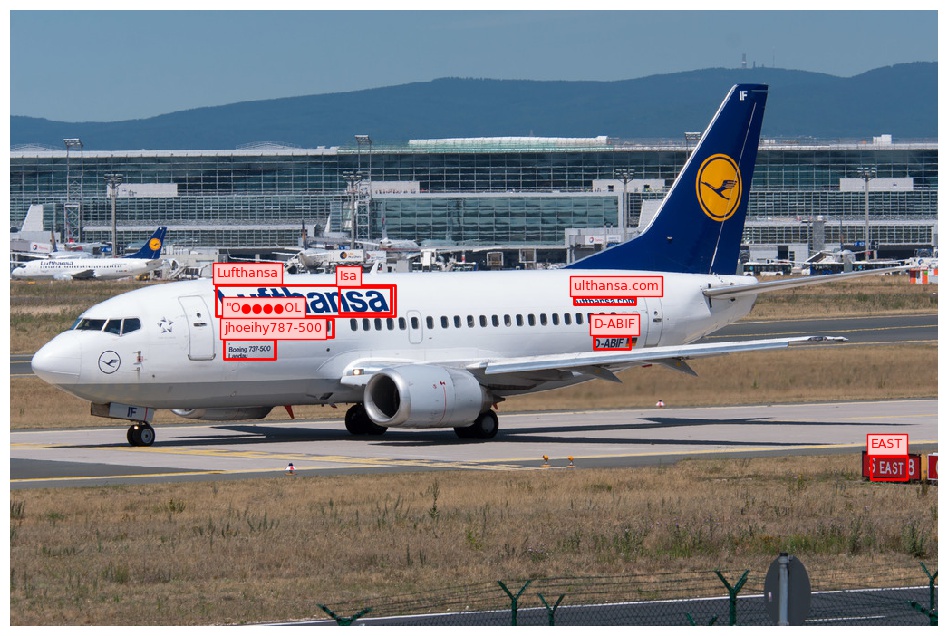} &
\includegraphics[width=\linewidth]{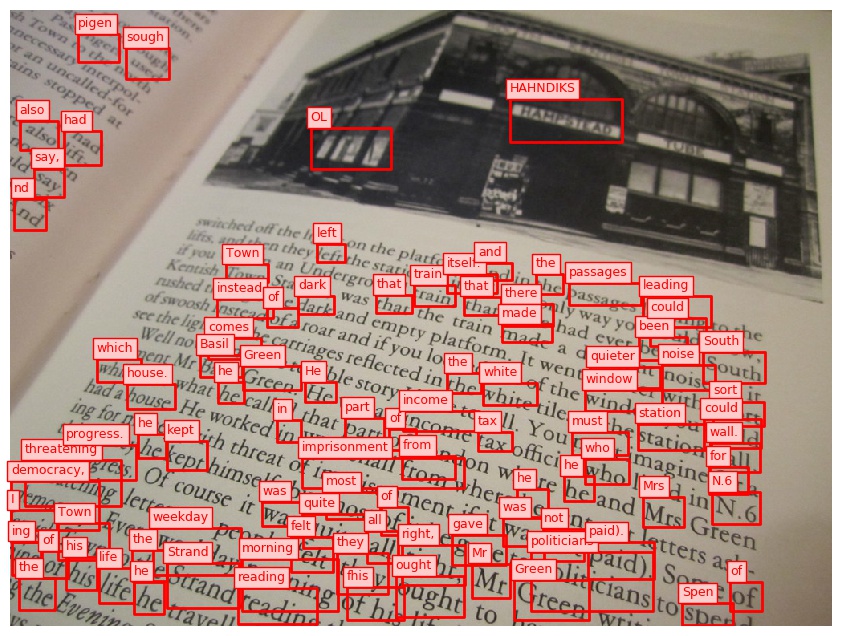} &
\includegraphics[width=\linewidth]{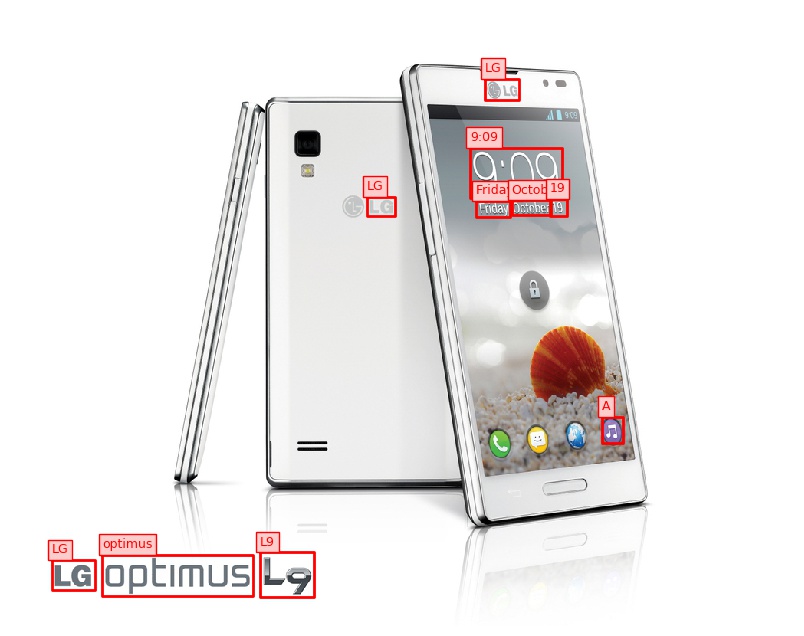} &
\includegraphics[width=\linewidth]{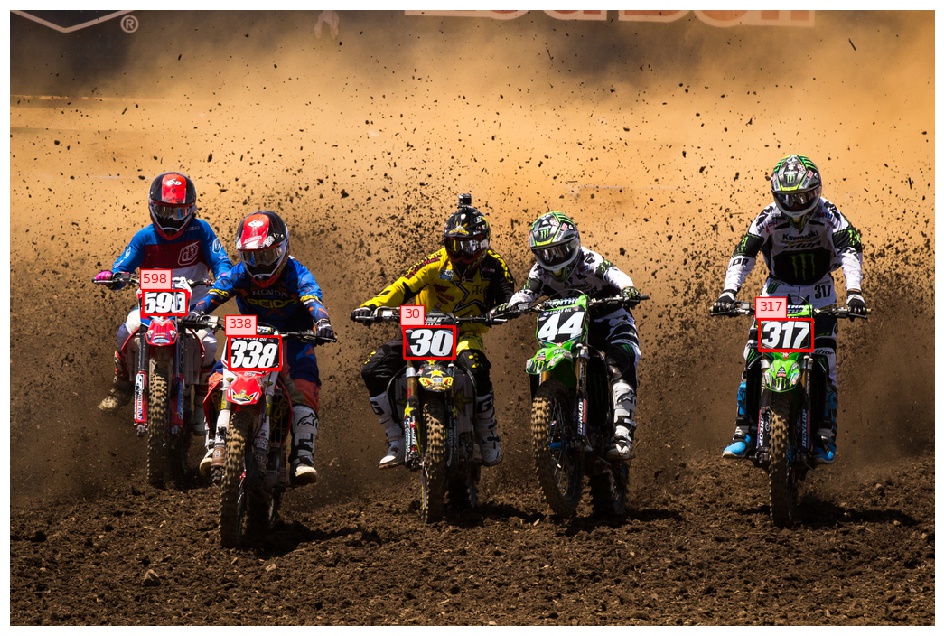} \\
(e) \textit{what airline is the plane from?} &
(f) \textit{what was mr. green's first name?} &
(g) \textit{what time is displayed on the phone's screen?} &
(h) \textit{what number is on the bike on the right?} \\
\methodname: \ocr{lufthansa} &
\methodname: \vocab{charles} &
\methodname: \ocr{9:09} &
\methodname: \vocab{30} \\
human: \textbf{lufthansa} &
human: \textbf{basil} &
human: \textbf{9:09}; \textbf{no} &
human: \textbf{317} \\
\end{tabular}
\vspace{0.5em}
\caption{Additional qualitative examples from our \methodname model on the TextVQA validation set. The red boxes show the OCR results (best viewed in 400\%; \ocr{orange} words from OCR tokens and \vocab{blue} words from fixed answer vocabulary).}
\label{fig:vis_supp_textvqa}
\end{figure*}

\begin{figure*}[t]
\vspace{\doublecoltopmargin}
\footnotesize
\centering
\begin{tabular}{p{0.22\linewidth}p{0.22\linewidth}p{0.22\linewidth}p{0.22\linewidth}}
\includegraphics[width=\linewidth]{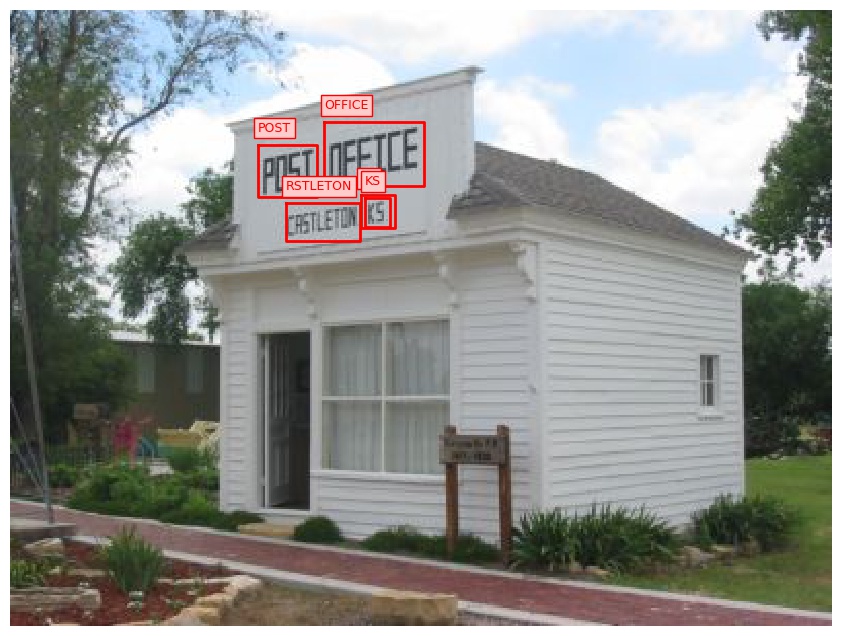} &
\includegraphics[width=\linewidth]{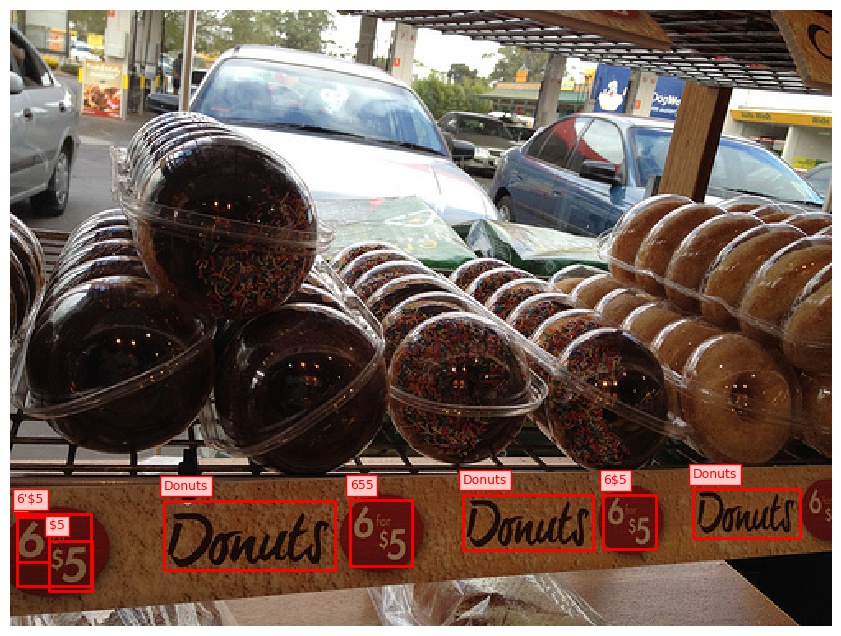} &
\includegraphics[width=\linewidth]{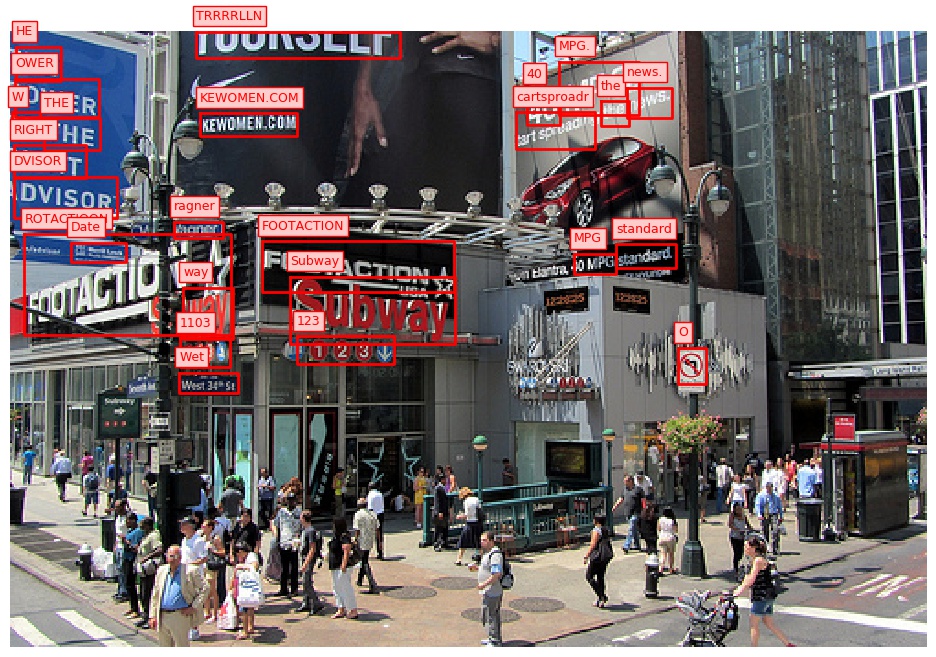} &
\includegraphics[width=\linewidth]{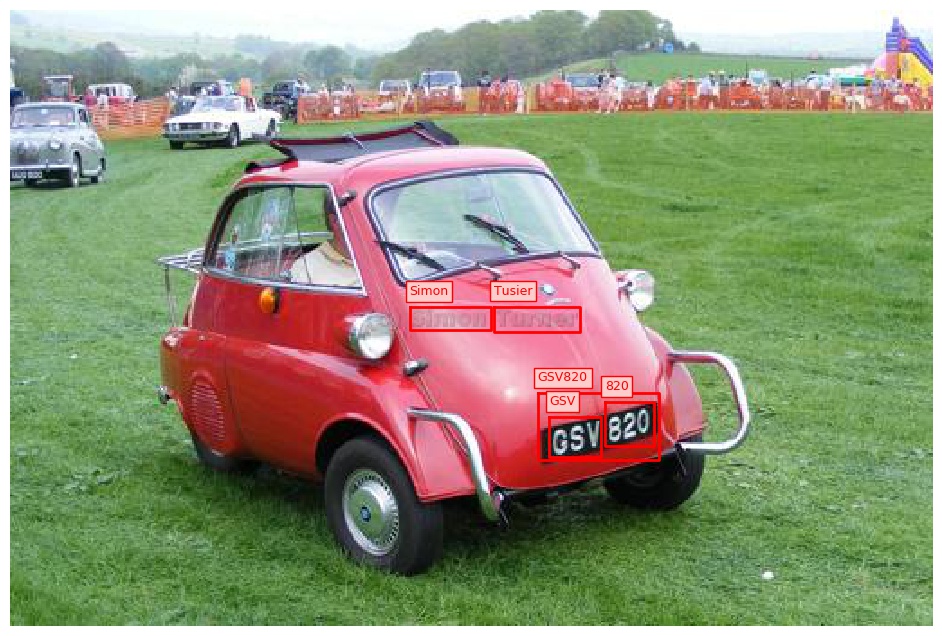} \\
(a) \textit{What is this building used for according to the sign above it?} &
(b) \textit{What can you get 6 of for \$5?} &
(c) \textit{where can I buy shoes here?} &
(d) \textit{What is the license plate number on the red car?} \\
\methodname: \ocr{post} \vocab{office} &
\methodname: \ocr{donuts} &
\methodname: \vocab{public} \vocab{market} &
\methodname: \ocr{gsv} \ocr{820} \\
GT: \textbf{post office} &
GT: \textbf{donuts} &
GT: \textbf{footaction} &
GT: \textbf{gsv 820} \\
\\
\includegraphics[width=\linewidth]{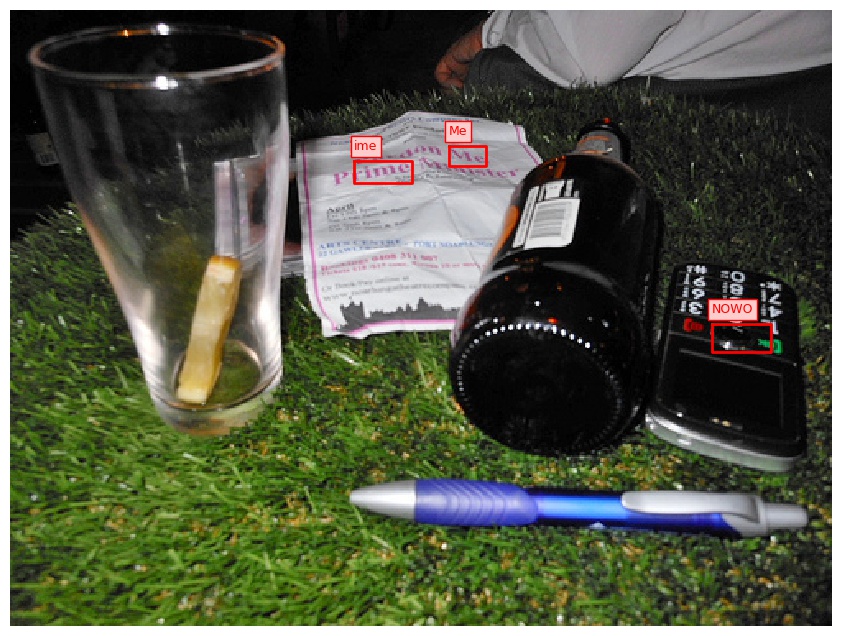} &
\includegraphics[width=\linewidth]{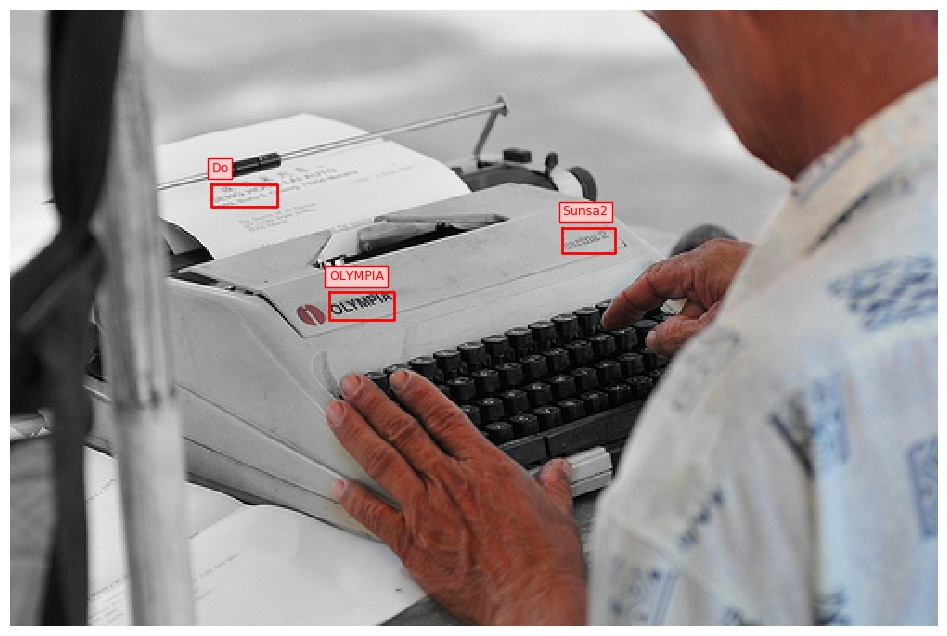} &
\includegraphics[width=\linewidth]{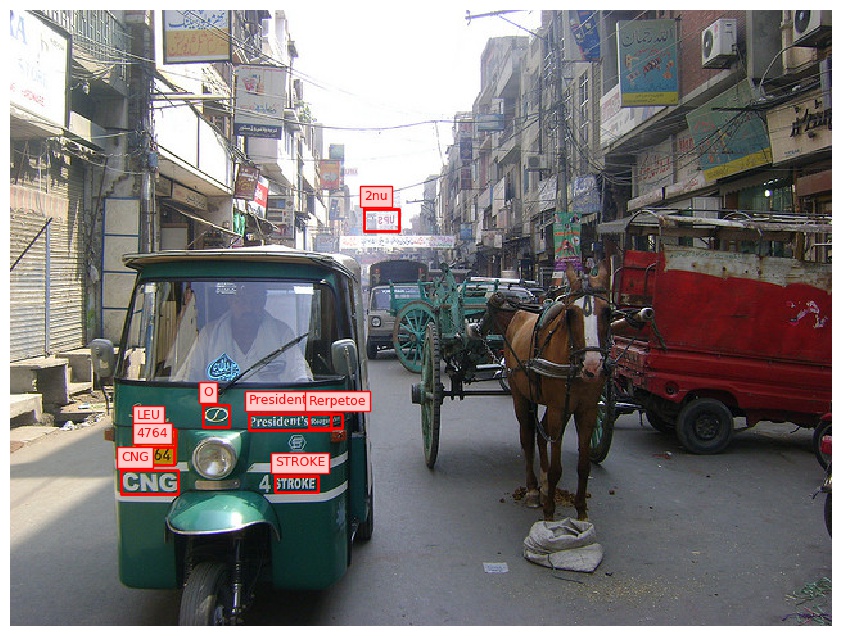} &
\includegraphics[width=\linewidth]{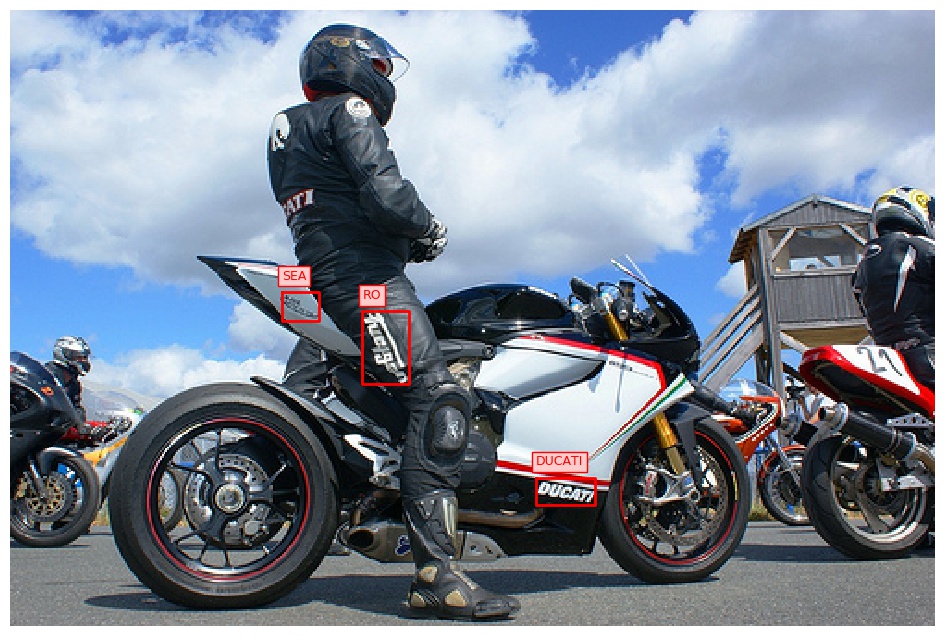} \\
(e) \textit{What does the large pink text say?} &
(f) \textit{What brand of typewriter is being used?} &
(g) \textit{What 4-digit number is on the yellow stick in front of the green car?} &
(h) \textit{What brand is the bike in front?} \\
\methodname: \ocr{me} &
\methodname: \ocr{olympia} &
\methodname: \ocr{4764} &
\methodname: \vocab{ducati} \\
GT: \textbf{pardon me prime minister} &
GT: \textbf{olympia} &
GT: \textbf{4764} &
GT: \textbf{ducati} \\
\end{tabular}
\vspace{0.5em}
\caption{Additional qualitative examples from our \methodname model on the ST-VQA validation set. The red boxes show the OCR results (best viewed in 400\%; \ocr{orange} words from OCR tokens and \vocab{blue} words from fixed answer vocabulary).}
\label{fig:vis_supp_stvqa}
\end{figure*}

\begin{figure*}[t]
\vspace{\doublecoltopmargin}
\footnotesize
\centering
\begin{tabular}{p{0.22\linewidth}p{0.22\linewidth}p{0.22\linewidth}p{0.22\linewidth}}
\includegraphics[width=\linewidth]{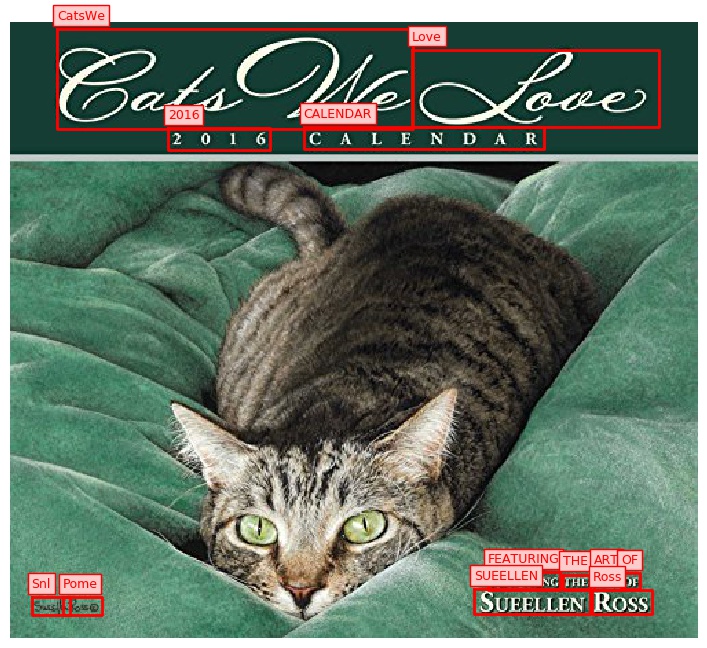} &
\includegraphics[width=\linewidth]{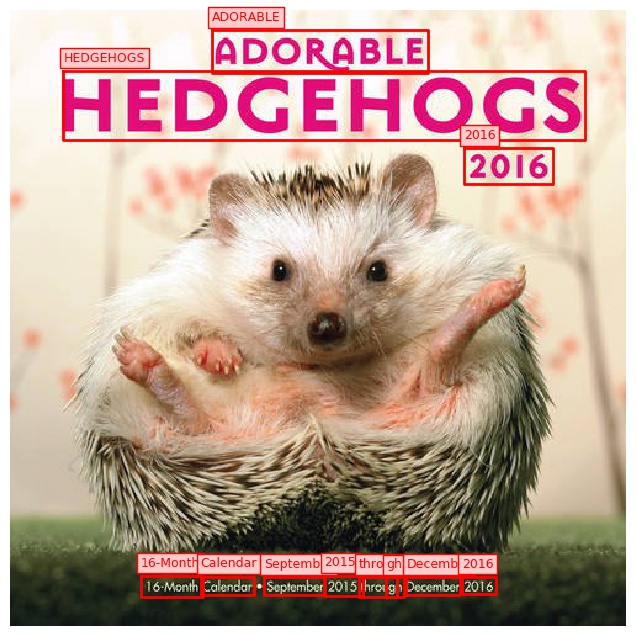} &
\includegraphics[width=\linewidth]{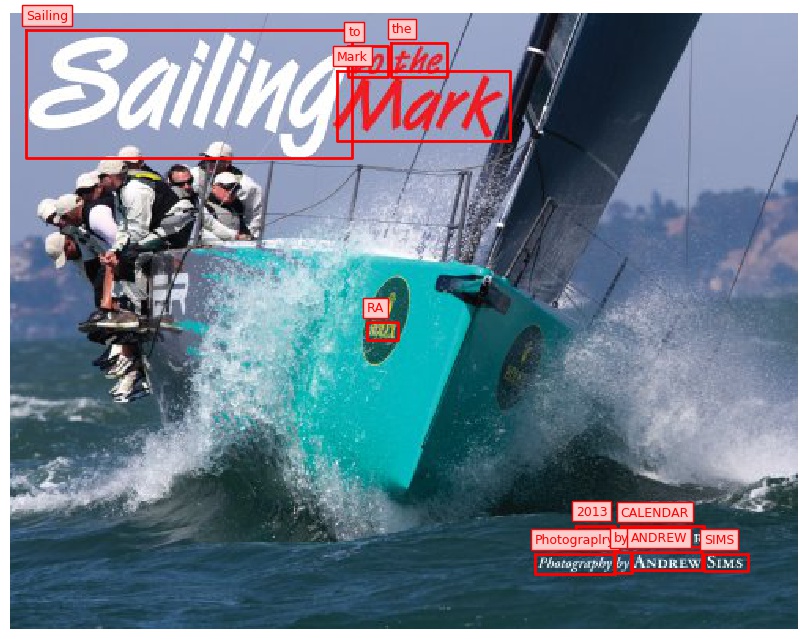} &
\includegraphics[width=\linewidth]{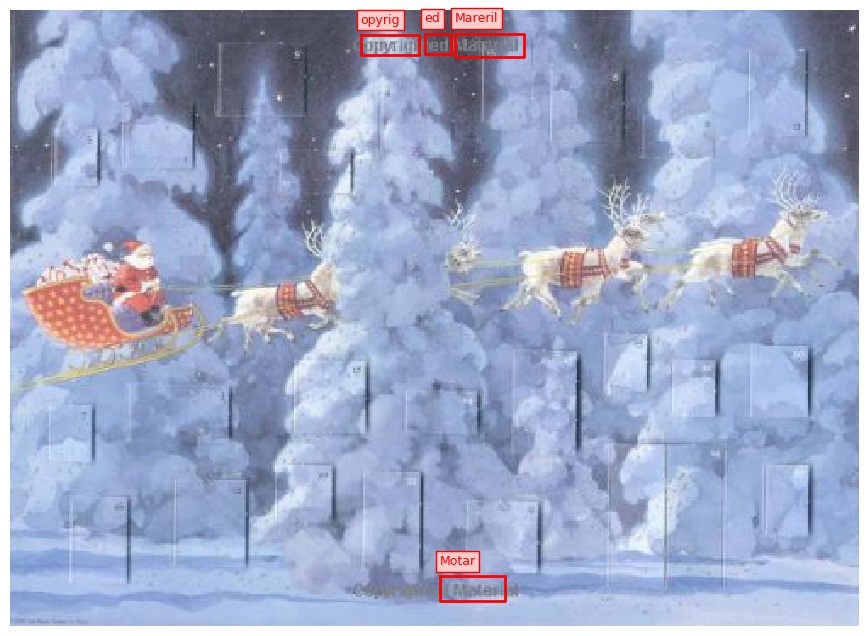} \\
(a) \textit{Who is the author of this book?} &
(b) \textit{Which year's calendar is this?} &
(c) \textit{What is the title of this book?} &
(d) \textit{What is the genre of this book?} \\
\methodname: \ocr{sueellen} \ocr{ross} &
\methodname: \vocab{2016} &
\methodname: \ocr{sailing} \ocr{to} \ocr{the} \ocr{mark} \ocr{2013} \ocr{calendar} &
\methodname: \vocab{arts} \vocab{\&} \vocab{photography} \\
GT: \textbf{sueellen ross} &
GT: \textbf{2016} &
GT: \textbf{sailing to the mark 2013 calendar} &
GT: \textbf{calendars} \\
\end{tabular}
\vspace{0.5em}
\caption{Additional qualitative examples from our \methodname model on the OCR-VQA validation set. The red boxes show the OCR results (best viewed in 400\%; \ocr{orange} words from OCR tokens and \vocab{blue} words from fixed answer vocabulary).}
\vspace{57em}
\label{fig:vis_supp_ocrvqa}
\end{figure*}

As mentioned in Sec.~4.1 in the main paper, we find that OCR failure is a major source of error for our \methodname model's predictions. Figure~\ref{fig:vis_supp_textvqa_ocrfail} shows cases on the TextVQA dataset where the OCR system fails to precisely localize the corresponding text tokens in the image, suggesting that our model's accuracy can be improved with better OCR systems.

Figure~\ref{fig:vis_supp_textvqa}, \ref{fig:vis_supp_stvqa}, and \ref{fig:vis_supp_ocrvqa} shows additional qualitative examples from our \methodname model on the TextVQA dataset, ST-VQA, and OCR-VQA datasets, respectively. While our model occasionally fails when reading a large piece of text or resolving the relation between text and objects as in Figure~\ref{fig:vis_supp_textvqa} (f) and (h), in most cases it learns to identify and copy text tokens from the image and combine them with its fixed vocabulary to predict an answer.

\end{document}